%% file: main.tex
\documentclass[10pt,twocolumn,letterpaper]{article}

\usepackage{iccv}      
\usepackage{colortbl} 
\usepackage{booktabs}
\usepackage{multirow}
\usepackage{bbding}
\usepackage{amsmath}
\usepackage{pifont}
\usepackage{bbm}


\definecolor{iccvblue}{rgb}{0.21,0.49,0.74}
\usepackage[pagebackref,breaklinks,colorlinks,allcolors=iccvblue]{hyperref}

\def\paperID{12592} 
\def\confName{ICCV}
\def\confYear{2025}

\title{Adaptive Dual Uncertainty Optimization: Boosting Monocular 3D Object Detection under Test-Time Shifts
}

\author{
Zixuan Hu$^{1}$ \quad Dongxiao Li$^{1}$ \quad Xinzhu Ma$^{3}$ \quad Shixiang Tang$^{3}$ \quad Xiaotong Li$^{1}$ \\
Wenhan Yang$^{2}$ \quad Ling-Yu Duan$^{1,2}$\thanks{Corresponding Author.} \\
\normalsize
$^{1}$School of Computer Science, Peking University, Beijing, China \\
\normalsize
$^{2}$Peng Cheng Laboratory, Shenzhen, China, $^{3}$The Chinese University of Hong Kong, Hongkong, China. \\
{\tt\small \{hzxuan, lingyu\}@pku.edu.cn
}
}

\begin{document}
\maketitle

\input{sec/0_abstract}    
\input{sec/1_intro}
\input{sec/2_related}
\input{sec/3_method}
\input{sec/4_experiment}
\input{sec/5_ablation}
\input{sec/6_conclusion}

{
    \small
    \bibliographystyle{ieeenat_fullname}
    \bibliography{main}
}
\input{sec/7_appendix}
\end{document}

%% file: sec/0_abstract.tex
\begin{abstract}
Accurate monocular 3D object detection (M3OD) is pivotal for safety-critical applications like autonomous driving, yet its reliability deteriorates significantly under real-world domain shifts caused by environmental or sensor variations. To address these shifts, Test-Time Adaptation (TTA) methods have emerged, enabling models to adapt to target distributions during inference. While prior TTA approaches recognize the positive correlation between low uncertainty and high generalization ability, they fail to address the dual uncertainty inherent to M3OD: semantic uncertainty (ambiguous class predictions) and geometric uncertainty (unstable spatial localization). To bridge this gap, we propose \textbf{D}ual \textbf{U}ncertainty \textbf{O}ptimization (\textbf{DUO}), the first TTA framework designed to jointly minimize both uncertainties for robust M3OD. Through a convex optimization lens, we introduce an innovative convex structure of the focal loss and further derive a novel unsupervised version, enabling label-agnostic uncertainty weighting and balanced learning for high-uncertainty objects. In parallel, we design a semantic-aware normal field constraint that preserves geometric coherence in regions with clear semantic cues, reducing uncertainty from the unstable 3D representation. This dual-branch mechanism forms a complementary loop: enhanced spatial perception improves semantic classification, and robust semantic predictions further refine spatial understanding. Extensive experiments demonstrate the superiority of DUO over existing methods across various datasets and domain shift types. The source code is available at \href{https://github.com/hzcar/DUO}{https://github.com/hzcar/DUO}.
\end{abstract}

%% file: sec/1_intro.tex
\begin{figure}[t]
\setlength{\abovecaptionskip}{-0.1cm} 
\begin{center}
\includegraphics[width=1.0\linewidth]{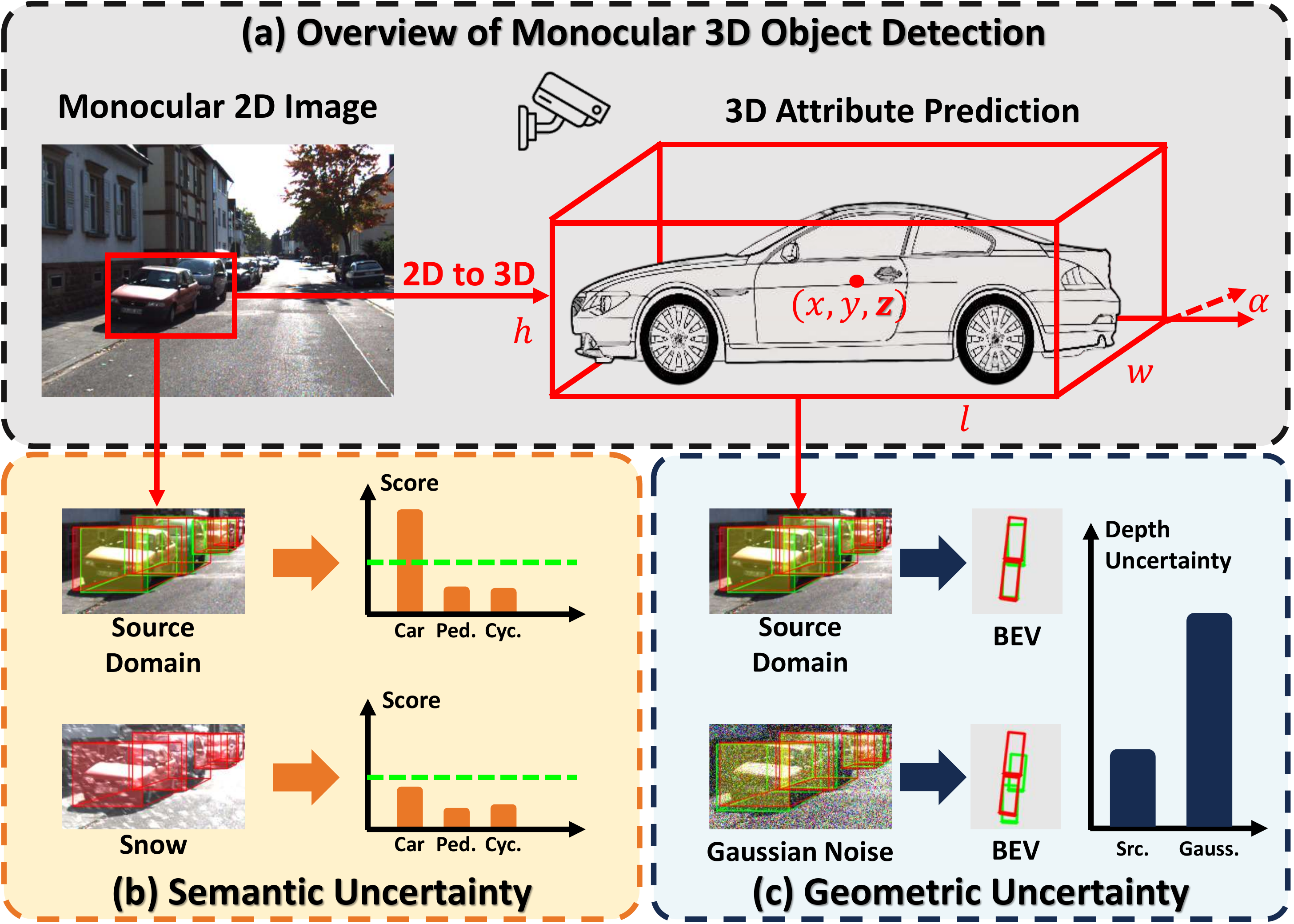}
\end{center}
\vspace{-0cm}
\caption{({\it a}) Illustration of M3OD.  ({\it b, c}) Under test-time shifts, increased semantic and geometric uncertainties lead to a degradation of semantic perception and spatial localization capabilities.}
\vspace{-0.6cm}
\label{fig:teaser}
\end{figure}

\section{Introduction}
\label{sec:intro}
Monocular 3D object detection (M3OD) \cite{m3od1,m3od2,m3od3} serves as a fundamental perception task, enabling agents to understand 3D scenes directly from 2D images, as shown in Fig. \ref{fig:teaser}(a). Due to its low cost and simple hardware configuration, M3OD has attracted widespread attention, leading to the development of numerous detectors~\cite{chen2020monopair,xu2023mononerd,luo2023latr}. However, when facing adverse weather conditions or sensor failures in real-world deployments, well-trained models often suffer severe performance degradation under data shifts~\cite{ben2010theory,hendrycks2018benchmarking,wang2020train}. Therefore, it is crucial to deal with the OOD generalization problem for M3OD.

To address the distribution shifting issue with minimal overhead, Test-Time Adaptation (TTA) has emerged as a critical paradigm, enabling source models to adapt to target distributions via online updates~\cite{iwasawa2021test,hu2025beyond,liang2024comprehensive}. The dominating strategy in this context involves minimizing prediction entropy, thereby reducing the uncertainty of the model on shifted data \cite{tent,hu2025beyond}. Despite the promising results, existing TTA approaches largely overlook the dual uncertainty inherent in 3D detection—semantic uncertainty (related to class predictions) \cite{Wang_2021_ICCV,S_2021_CVPR} and geometric uncertainty (related to spatial location) \cite{lu2021geometry,poggi2020uncertainty}, which is a significant difference compared to conventional 2D tasks.

To investigate these two types of uncertainty under test-time shifts, we analyze detection outcomes for objects subject to common real-world variations. As shown in Fig. \ref{fig:teaser} (b)\&(c), our empirical study reveals that both uncertainties increase markedly with data shifts, creating compounded error accumulation in 3D detection. Moreover, we find that existing uncertainty optimization techniques exhibit significant limitations: 
1)  Low-score object neglect. Entropy minimization fails to provide effective supervision for challenging objects with low detection scores, resulting in inevitable omissions. 
2) Spatial perception collapse. Direct minimization of depth uncertainty can cause model collapse, compromising the perception capacity of spatial attributes. 

In this work, to overcome the above limitations, we propose \textbf{D}ual \textbf{U}ncertainty \textbf{O}ptimization (\textbf{DUO}), the first TTA  framework for joint semantic-geometric uncertainty minimization.
Specifically, through the lens of convex optimization theory \cite{bertsekas2003convex,touchette2005legendre}, we present an Legendre–Fenche structure of the focal loss \cite{lin2017focal} and reconstruct the semantic uncertainty minimization as a dual optimization problem.
Building on this foundation, we apply higher-order approximation analysis to derive a novel Conjugate Focal Loss. This loss breaks the label-dependence barrier in the original objective and introduces a dynamic weighting mechanism for balanced training, effectively reducing the omissions of objects with low scores.
In parallel, we introduce a normal field constraint that enforces local consistency of surface normals in regions with high semantic confidence. This spatial coherence clarifies geometric cues, thereby reducing geometric uncertainty.
Together, this dual-branch design creates a complementary loop where semantic confident regions bootstrap geometric feature learning,  while enhanced spatial perception guides semantic refinement.

We evaluate the effectiveness and generalizability of our method through experiments on the KITTI dataset \cite{geiger2012we} with 13 corruption shift types, achieving state-of-the-art results with average improvements of \textbf{+2.2} ${\rm AP}_{{\rm 3D}|{\rm R}_{40}}$ in the Car category. Furthermore, we showcase its superior performance in addressing real-world shift scenarios (daytime $\leftrightarrow$ night, sunny $\leftrightarrow$ rainy) of nuScenes dataset \cite{caesar2020nuscenes}, yielding an average gain of $\textbf{+18\%}$ compared to existing methods.

\textbf{Contributions:} 
1) To the best of our knowledge, we pioneer dual uncertainty optimization in M3OD by establishing the first TTA framework that jointly minimizes semantic and geometric uncertainties, addressing a critical reliability gap in real-world deployments.
2) Through the lens of convex optimization theory, we derive a novel Conjugate Focal Loss that enables label-agnostic uncertainty weighting and balanced learning for low-score objects. This approach is inherently compatible with the source phase, requiring no additional hyperparameter tuning.
3) We introduce a normal field constraint that enforces the stability of geometric representation with semantic guidance, resolving ambiguous spatial predictions.
4) We analyze and verify that our dual-branch design creates a complementary loop of two types of uncertainty optimization, resulting in significantly improved performance over existing TTA methods.

%% file: sec/2_related.tex
\vspace{-2mm}
\section{Related Work}
\vspace{-1mm}
\label{sec:relatedwork}
\noindent\textbf{Monocular 3D Object Detection (M3OD)} aims to perceive 3D objects from a single 2D image. Existing methods in M3OD can be broadly categorized based on their use of extra data sources, such as CAD models~\cite{liu2021autoshape}, dense depth maps~\cite{ding2020learning,wang2021depth}, or LiDAR~\cite{reading2021categorical,huang2022monodtr}. In this paper, we focus exclusively on approaches that utilize only monocular images, due to their computational efficiency and lower deployment costs. Previous studies, such as MonoDLE~\cite{Ma_2021_CVPR} and PGD~\cite{wang2022probabilistic}, have identified depth estimation as a critical bottleneck in M3OD. To address this, many works leverage multiple geometric cues to integrate diverse depth predictions. For example,
MonoFlex~\cite{zhang2021objects} integrates depth prediction by combining direct regression with multi-keypoint estimation; MonoGround~\cite{qin2022monoground} incorporates the ground plane as prior information; 
and MonoCD \cite{yan2024monocd} exploits the complementary properties of multi-head estimation. In this paper, we investigate how to enhance the detection performance of such detectors under test-time shifts.

\noindent\textbf{Test-Time Adaptation (TTA)} aims to enhance model performance on out-of-distribution samples during inference. Depending on whether the source training process is modified, existing TTA methods can be mainly divided into two groups: Test-Time Training \cite{ttt,hakim2023clust3,liu2024depth} and Fully Test-Time Adaptation \cite{boudiaf2022parameter,zhao2023delta,hu2025seva}.  In this paper, we focus on Fully TTA, which adapts models without source data. Prior studies have demonstrated that reducing prediction uncertainty is an effective strategy for improving model generalization in various tasks \cite{lin2023video,gaofast,karmanov2024efficient,wang2024backpropagation}. These works have developed various strategies to model and optimize uncertainty. For instance, SAR \cite{sar} uses
entropy as a measure of classification uncertainty along with sharpness-aware optimization. DeYO \cite{deyo} incorporates entropy with a disentangled factor for uncertainty modeling. ReCAP \cite{hu2025beyond} models regional uncertainty, enforcing implicit data scaling for uncertainty optimization. MonoTTA \cite{lin2025monotta} optimizes positive and negative class uncertainties separately. However, unlike conventional 2D tasks, M3OD inherently exhibits both semantic and geometric uncertainties, which remain largely unexplored and lack an effective training paradigm. In this work, we explore the joint optimization of this dual uncertainty, enhancing the robustness of M3OD models.

%% file: sec/3_method.tex
\begin{figure*}[!t]
\begin{center}
\includegraphics[width=1.0\textwidth]{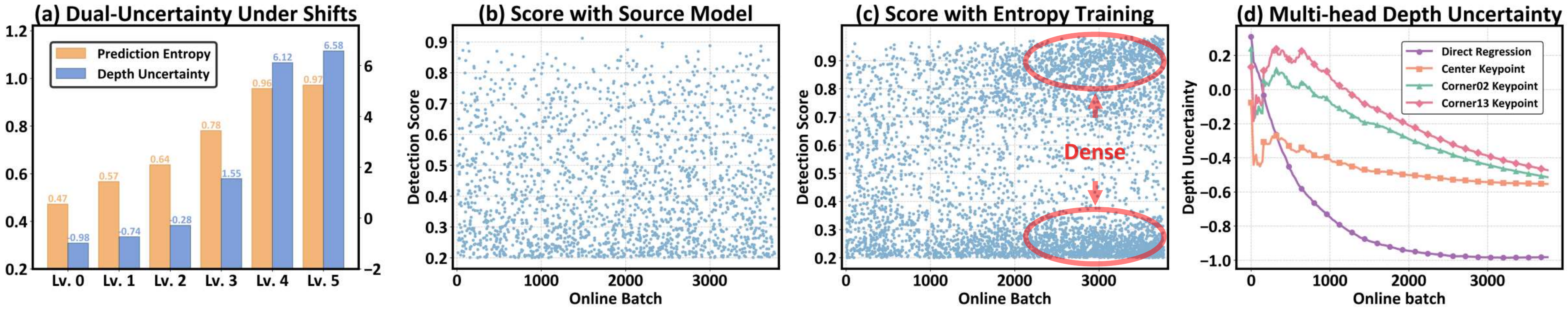}
\end{center}
\vspace{-0.6cm}
\caption{\label{fig:observe} Semantic uncertainty (prediction entropy) and geometric uncertainty (average depth uncertainty) of the M3OD model under domain shifts. ({\it a}) shows variations in dual-uncertainty as the shift level increases. ({\it b}) records detection scores of predicted objects during inference with the source model. ({\it c}) records scores during the entropy minimization process. ({\it d}) investigates the depth uncertainty of different heads during the optimization with uncertainty regression loss. Experiments are conducted on the Gauss. shift type of KITTI-C.}
\vspace{-0.3cm}
\end{figure*}
\section{Preliminary}
\noindent\textbf{Task Definition.} M3OD aims to predict the 3D and semantic attributes of objects from a single RGB image. Given an input image $\text{I} \in \mathbb{R}^{H \times W \times 3}$, the goal is to generate accurate 3D bounding boxes $
\{\mathcal{B}_i\}_{i=1}^{N}$ and semantic labels $\{\mathcal{C}_i\}_{i=1}^{N}$ for objects present in the scene, where $N$ denotes the number of objects. Each bounding box is typically parameterized as
$\mathcal{B}_i = (\mathbf{P}_i,\, \mathbf{D}_i,\, \mathbf{O}_i)$,
where $\mathbf{P}_i \in \mathbb{R}^3$ denotes the 3D center position, $\mathbf{D}_i \in \mathbb{R}^3$ represents the shape dimensions, and $\mathbf{O}_i \in [-\pi, \pi]$ encodes the orientation. The multi-task nature of M3OD, which involves simultaneous estimation of both geometric and semantic attributes, poses significant challenges for achieving precise and coherent predictions.

\begin{figure*}[!t]
\begin{center}
\includegraphics[width=1.0\textwidth]{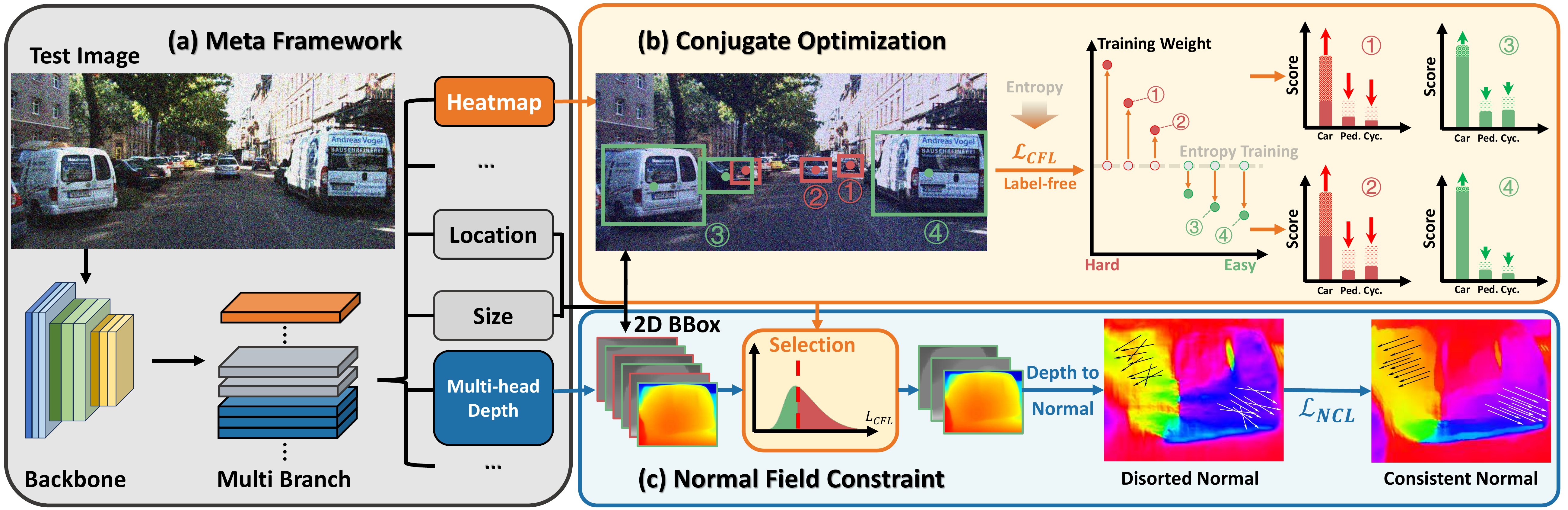}
\end{center}
\vspace{-0.6cm}
 \caption{Overview of our DUO method. ({\it a}) Meta-framework of the base models, where a backbone network connects to multi-branch predictors for estimating various object properties. ({\it b}) Through our theoretical conjugate optimization framework, we derive a conjugate focal loss $\mathcal{L}_\text{CFL}$ that can adaptively adjust training weights across all classes without the ground-truth labels, while also identifying low-semantic-uncertainty regions for geometric constraints (Sec. \ref{sec:5.1}). ({\it c}) DUO employs the efficient Sobel operator to convert the depth map to a normal field and design a normal consistency constraint $\mathcal{L}_\text{NCL}$ to enhance geometric consistency in selected regions (Sec. \ref{sec:5.2}).}
\label{fig:pipeline}
\vspace{-0.45cm}
\end{figure*}

\noindent
{\bf Meta-framework.} As shown in Fig. \ref{fig:pipeline}(a), M3OD models typically employ a backbone network connected with multiple branches to predict various properties, \eg, score heatmaps, depth maps, \etc. Since depth estimation is widely recognized as a key bottleneck \cite{Ma_2021_CVPR,wang2022probabilistic}, many approaches employ a multi-head depth estimator to improve prediction accuracy. This estimator comprises a direct regression depth head and multiple geometric depth heads, each providing an individual depth prediction along with an associated uncertainty value. These outputs are then fused via an uncertainty-weighted average to produce the final depth prediction. We utilize the average uncertainty across all heads as our depth uncertainty metric. A detailed explanation is provided in the Appendix. \ref{sm:model}.

\noindent\textbf{TTA Setting.} TTA addresses the challenge of distribution shifts by enabling a pre-trained model to adapt to the target distribution during inference, without the need for labeled data. Unlike traditional training on fixed datasets, TTA operates in an online manner, where the model $h_{\theta}$ with parameter set $\theta$ produces detection outputs while concurrently updating its parameters based on incoming test data:
\vspace{-0.1cm}
\begin{equation}
    \{\mathcal{B}_t\}, \{\mathcal{C}_t\}\leftarrow h_{\theta}(\text{I}_t),\; \theta\leftarrow\theta-\nabla\mathcal{L}_{tta}(\text{I}_t),
\label{eq:tta}
\end{equation}
where $\text{I}_t$ denotes the incoming test data and $\mathcal{L}_{tta}$ denotes the loss function used for self-training during adaptation.

\vspace{-0.1cm}
\section{Uncovering Dual-Uncertainty under Shifts}
\label{sec:observation}
Since TENT \cite{tent} revealed the positive correlation between prediction uncertainty and generalization error under distribution shifts, numerous TTA methods have emerged to optimize uncertainty metrics. While previous works have developed diverse measures for conventional 2D tasks \cite{sar,deyo}, the compound semantic and geometric uncertainties in 3D perception remains critically unexplored in the TTA field. In this work, we investigate both types of uncertainty within the context of M3OD to understand their distinct roles.

Specifically, to quantify the variation of uncertainty under distribution shifts, we track two metrics: semantic prediction entropy and geometric depth uncertainty (average uncertainty of multi-head depth estimator). As shown in Fig. \ref{fig:observe}(a), both metrics demonstrate a consistent upward trend as distribution shifts intensify, indicating that more severe shifts lead to higher model uncertainty. Furthermore, we analyze the independent optimization effects of two uncertainties and make the following key observations:

\noindent {\bf Observation 1}:  \emph{Conventional entropy minimization exacerbates imbalance distribution of detection scores.} Different from classification scenarios, object detection suffers from extreme foreground-background imbalance, which hinders the effective optimization of hard positive objects \cite{lin2017focal,chen2020foreground}. As shown in Fig. \ref{fig:observe}(b)\&(c), entropy minimization yields a marginal gain for low-score objects while significantly boosting high-score predictions, further exacerbating the imbalance and leading to omissions of low-score objects.

\noindent {\bf Observation 2}: \emph{Minimizing depth uncertainty directly causes the model collapse of the multi-head depth estimator.} To minimize depth uncertainty, we apply the uncertainty regression loss across multiple depth heads (detailed in the Appendix. \ref{sm:model}). As shown in Fig. \ref{fig:observe}(d), the regression head which lacks any geometric constraints, exhibits a significantly faster decline of uncertainty compared to keypoint heads. This rapid convergence reduces the multi-head system to a single deterministic head, undermining the model's ability to perform robust spatial understanding.

\vspace{-0.15cm}

\section{Methodology}
Based on the above observations, we propose a TTA method \textbf{D}ual \textbf{U}ncertainty \textbf{O}ptimization (\textbf{DUO}) for M3OD, which mainly leverages two novel designs, \ie, conjugate optimization framework and normal consistency constraint, to compatibly optimize semantic and geometric uncertainties.

\subsection{Semantic Uncertainty Conjugate Optimization}
\label{sec:5.1}
To address the imbalance problem, it is crucial to devote more attention to challenging objects with high uncertainty. A straightforward approach is to employ the focal loss \cite{lin2017focal} which increases the weight of predictions with low probability. However, focal loss cannot assign an appropriate weight without the ground-truth label and fails to adapt effectively in unsupervised settings. To provide a more flexible weighting solution, we leverage convex optimization theory \cite{bertsekas2003convex,touchette2005legendre} (a classical analytical tool) to explore the focal loss via a Legendre–Fenchel structure. Building on this foundation, we perform a higher-order approximation analysis to derive a novel loss that can dynamically adjust the weighting without relying on labels, as shown in Fig. \ref{fig:pipeline}(b).

\noindent\textbf{Legendre–Fenchel Structure.} Given the source model $h_\theta$ pre-trained with the focal loss, we denote $h\triangleq h_\theta(x)$ and its corresponding loss function can be formulated as:
\begin{equation}
\begin{aligned}
    \mathcal{L}_{\text{FL}}(x, y) =-\alpha (1-p)^\gamma&\cdot y\log p,
\label{eq:focal loss}
\end{aligned}
\end{equation}
where $p\triangleq \texttt{softmax}(h)$ is the normalized probability over the classes, $x$ is a detected object, and $y$ is the one-hot coding of the ground-truth label. $\alpha$ and $\gamma$ are hyperparameters controlling the weighting of uncertain predictions. Motivated by \cite{goyal2022test}, we reformulate it as the following structure: 
\begin{equation}
\begin{aligned}
    \mathcal{L}_{\text{FL}} &= f(h)-y^\top g(h),\\
     f(h)=\alpha\log s,\, g(h&)= \alpha h+\alpha((1-p)^\gamma -1) \log p,
\end{aligned}                
\end{equation}
where $s \triangleq e^{h_1}+...+e^{h_\text{c}}$ is the sum over the exponential outputs of the model and $\text{c}$ is the number of classes. Refer to the Appendix. \ref{sm:proof} for missing proofs of this subsection.

\noindent\textbf{Problem Reconstruction.} Under this structure, the optimization problem can be regarded as finding an optimal representation $h$ that minimizes the empirical loss. According to the Legendre–Fenchel condition \cite{touchette2005legendre}, the existence of the conjugate function $f^{*}$ is equivalent to the invertibility of the function $g$. This critical condition is formally established in our work (see Appendix. \ref{sm:proof}), serving as a fundamental theoretical guarantee for our method. Therefore, the minimum value of the objective can be expressed as follows:
\begin{equation}
\small
    \min_h \{f(h)-y^\top g(h)\} = \min_{z=g(h)} \{f\circ g^{-1}(z)-y^\top z\}=f^{*}(y).
\label{eq:conjugate function}
\end{equation}
Building on the common assumption that the representation $h$ pre-trained from the large source dataset is already close to a locally optimal solution $h_0$ \cite{erhan2010does,mori2022power}, we can convert the problem into the following relationships:
\begin{equation} 
f \circ g^{-1}(z) - y^{\top} z = f^*(y), \quad\nabla_z (f \circ g^{-1}) = y.  
\label{conjugate relation} 
\end{equation}
 
\noindent\textbf{Conjugate Focal Loss.} 
By applying the chain rule of gradient computation and higher-order approximation, we can obtain the following estimation:
\begin{equation}
\begin{aligned}
\small
    &y_0\triangleq \frac{\nabla_h (f\circ g^{-1})}{\nabla_h z}|_{z=g(h)}=\nabla_h g(h)^{-1}\cdot\nabla_h f(h) \\
    &\approx (I+\gamma(1-\log p)\cdot pp^\top-\gamma \log p \cdot \text{diag}(p)))^{-1}p,
\end{aligned}
\end{equation}
where $\text{diag}(\cdot)$ denotes the diagonal matrix and $I$ denotes the identity matrix. This estimation $y_0$ ensures that the loss aligns closely with the conjugate function. Ultimately, we substitute it into Equ. \ref{conjugate relation}, yielding an unsupervised approximation of the conjugate function:
\begin{equation}
\begin{aligned}
\mathcal{L}_{\text{CFL}}(x)= & -\alpha (1-p)^\gamma (I+\gamma(1-\log p)\cdot p^{\top} p \\
 -&\gamma \log p \cdot \text{diag}(p))^{-1} p \log p.
\end{aligned}
\label{eq:CF loss}
\end{equation}
In contrast to the focal loss in Equ. \ref{eq:focal loss}, our derived \textbf{C}onjugate \textbf{F}ocal \textbf{L}oss (CFL) offers a dynamic adjustment of uncertainty weighting without relying on ground-truth labels and it confers the following advantages:

\noindent\textit{Static vs. Dynamic Adjustment.} 
The vanilla focal loss uses a fixed weighting term for the ground-truth class to focus on high-uncertainty predictions. In contrast, our CFL not only incorporates $(1-p)^\gamma$ to address the imbalance training issue but also dynamically adjusts the weighting across all classes based on the term $(I+\gamma(1-\log p) pp^\top -\gamma \log p \cdot \text{diag}(p))^{-1}$, which encodes inter-class prediction relationships.

\noindent\textit{Ground-Truth Independence.} While focal loss requires labels to compute the weight for the loss term, CFL operates solely based on the prediction probability, eliminating the need for labeled data during TTA.

\noindent\textit{Compatible Hyperparameters with Source Phase.} Our theoretical analysis suggests that hyperparameters $\alpha, \gamma$ should remain consistent with their values used in focal loss during the source training. This compatibility provides a practical advantage for TTA scenarios, eliminating the need for extensive hyperparameter tuning. The validity of this consistent setting is empirically verified in Appendix. \ref{sm:ablation}.

\begin{table*}[t]
\centering
\vspace{-3.5mm}
\caption{Comparisons with state-of-the-art methods on the KITTI-C \emph{validation} set (severity level 5) with MonoFlex as the base model. We highlight the best and second results with {\bf bold} and \underline{underline} respectively.}
\vspace{-0.1cm}
\resizebox{0.9\textwidth}{!}{
{
\fontsize{20}{25}\selectfont
\begin{tabular}{c|c|ccc|ccc|cccc|ccc|c}
\toprule[1pt]
\multicolumn{16}{c}{Car Category}\\
\toprule[1pt]
\multirow{2}{*}{\centering\textbf{Method}}  & \multirow{2}{*}{\centering\textbf{Reference}}   & \multicolumn{3}{c|}{Noise}              & \multicolumn{3}{c|}{Blur}                          & \multicolumn{4}{c|}{Weather}                       & \multicolumn{3}{c|}{Digital}                       &     \multirow{2}{*}{\centering\textbf{Avg}}       \\
 &  & Gauss.       & Shot       & Impul.     & Defoc.     & Glass      & Motion      & Snow       & Frost      & Fog        & Brit.      & Contr.    & Pixel       & Sat.         \\
\toprule[1pt]
\multicolumn{1}{c|}{MonoFlex}                                &       CVPR'21
& 0.00   & 0.00  & 0.00   & 0.00   & 11.02 & 0.00   & 5.07  & 12.94 & 7.70  & 14.46 & 0.00   & 2.13  & 5.65  & 4.54\\
\multicolumn{1}{l|}{\quad$\bullet$ TENT}                                &       ICLR'21
& 5.31   & 11.09 & \underline{6.40}   & 2.22   & 25.49 & 2.14   & 23.88 & 28.96 & 35.40 & 37.07 & 24.67  & 22.59 & 30.63 & 19.68\\
\multicolumn{1}{l|}{\quad$\bullet$ EATA}                                &       ICML'22
&5.44   & 12.12 & 4.67   & 2.75   & 25.66 & 2.45   & 24.77 & 28.99 & 35.67 & 36.95 & 24.33  & 22.47 & 34.10 & 20.03\\ 
\multicolumn{1}{l|}{\quad$\bullet$ DeYO}                                &       ICLR'24
&5.78   & 12.52 & 4.52   & 3.01   & 26.05 & 2.98   & 24.91 & \underline{29.40} & 35.19 & \underline{37.59} & 23.75  & 23.82 & \underline{34.33} & 20.30\\
\multicolumn{1}{l|}{\quad$\bullet$ MonoTTA}                                &       ECCV'24
&\underline{5.93}   & \underline{13.34} & 4.05   & \underline{3.35}   & \underline{28.10} & \underline{3.21}   & \underline{25.86} & 29.10 & \underline{36.43} & 37.18 & \underline{25.90}  & \underline{25.01} & 33.89 & \underline{20.87}\\ 

\rowcolor[HTML]{E6F1FF}
\multicolumn{1}{l|}{\quad$\bullet$ Ours}                                &       This paper
&\textbf{7.30} & \textbf{15.40} & \textbf{9.36} & \textbf{4.34} & \textbf{30.23} & \textbf{6.89} & \textbf{29.09} & \textbf{29.76} & \textbf{38.38} & \textbf{37.72} & \textbf{29.35} & \textbf{25.88} & \textbf{34.97} & \textbf{22.97}\\

\toprule[1pt]
\multicolumn{16}{c}{Pedestrian Category}\\
\toprule[1pt]
\multirow{2}{*}{\centering\textbf{Method}}  & \multirow{2}{*}{\centering\textbf{Reference}}   & \multicolumn{3}{c|}{Noise}              & \multicolumn{3}{c|}{Blur}                          & \multicolumn{4}{c|}{Weather}                       & \multicolumn{3}{c|}{Digital}                       &     \multirow{2}{*}{\centering\textbf{Avg}}       \\
 &  & Gauss.       & Shot       & Impul.     & Defoc.     & Glass      & Motion      & Snow       & Frost      & Fog        & Brit.      & Contr.    & Pixel       & Sat.         \\
\toprule[1pt]
\multicolumn{1}{c|}{MonoFlex}                                &       CVPR'21
&0.00   & 0.00  & 0.00   & 0.00   & 4.09  & 0.00   & 0.51  & 1.67  & 1.77  & 2.09  & 0.00   & 0.32  & 0.95  & 0.88\\
\multicolumn{1}{l|}{\quad$\bullet$ TENT}                                &       ICLR'21
&0.71   & 2.07  & 0.86   & 1.36   & \underline{12.70} & 0.81   & 6.72  & 8.33  & 11.89 & 12.79 & 7.42   & 6.45  & 9.78  & 6.30\\ 
\multicolumn{1}{l|}{\quad$\bullet$ EATA}                                &       ICML'22
&0.98   & 2.15  & 0.76   & 1.44   & 12.23 & 0.80   & 6.77  & 8.67  & 11.60 & 13.62 & 7.53   & 6.47  & 10.33 & 6.41\\ 
\multicolumn{1}{l|}{\quad$\bullet$ DeYO}                                &       ICLR'24
&1.03   & 2.25  & \underline{0.91}   & 1.64   & 11.85 & 0.80   & 6.63  & \textbf{9.09}  & 11.77 & \textbf{13.99} & 7.59   & 6.39  & 10.57 & 6.50\\
\multicolumn{1}{l|}{\quad$\bullet$ MonoTTA}                                &       ECCV'24
&\underline{1.77}   & \underline{2.88}  & 0.34   & \underline{1.78}   & 12.38 & \underline{0.82}   & \underline{7.03}  & 9.02  & \underline{12.31} & 13.11 & \underline{7.75}   & \underline{7.10}  & \underline{11.08} & \underline{6.72}\\ 

\rowcolor[HTML]{E6F1FF}
\multicolumn{1}{l|}{\quad$\bullet$ Ours}                                &      This paper
&\textbf{1.89} & \textbf{3.08}  & \textbf{1.54} & \textbf{1.86} & \textbf{13.53} & \textbf{1.75} & \textbf{7.68}  & \textbf{9.09}  & \textbf{12.66} & \textbf{13.99} & \textbf{7.81}  & \textbf{7.27}  & \textbf{11.27} & \textbf{7.19}\\

\toprule[1pt]
\multicolumn{16}{c}{Cyclist Category}\\
\toprule[1pt]
\multirow{2}{*}{\centering\textbf{Method}}  & \multirow{2}{*}{\centering\textbf{Reference}}   & \multicolumn{3}{c|}{Noise}              & \multicolumn{3}{c|}{Blur}                          & \multicolumn{4}{c|}{Weather}                       & \multicolumn{3}{c|}{Digital}                       &     \multirow{2}{*}{\centering\textbf{Avg}}       \\
 &  & Gauss.       & Shot       & Impul.     & Defoc.     & Glass      & Motion      & Snow       & Frost      & Fog        & Brit.      & Contr.    & Pixel       & Sat.         \\
\toprule[1pt]
\multicolumn{1}{c|}{MonoFlex}                                &       CVPR'21
&0.00   & 0.00  & 0.00   & 0.00   & 0.24  & 0.00   & 2.14  & 2.33  & 1.72  & 4.41  & 0.00   & 0.00  & 0.00  & 0.83\\ 
\multicolumn{1}{l|}{\quad$\bullet$ TENT}                                &       ICLR'21
&\underline{0.06}   & 0.14  & \underline{0.04}   & \underline{0.04}   & 4.55  & 0.93   & 6.63  & 8.23  & 11.94 & \underline{15.16} & \underline{7.72}   & 1.85  & 2.81  & 4.62\\ 
\multicolumn{1}{l|}{\quad$\bullet$ EATA}                                &       ICML'22
&0.05   & 0.15  & 0.03   & 0.02   & 4.66  & 1.10   & 6.73  & 7.58  & \underline{13.77} & 14.99 & 7.32   & 2.03  & 2.82  & 4.71\\ 
\multicolumn{1}{l|}{\quad$\bullet$ DeYO}                                &       ICLR'24
&\underline{0.06}   & \underline{0.19}  & 0.03   & 0.03   & \underline{4.91}  & 1.08   & 6.48  & 6.91  & \textbf{13.94} & 14.67 & 7.57   & 1.79  & 2.82  & 4.65\\
\multicolumn{1}{l|}{\quad$\bullet$ MonoTTA}                                &       ECCV'24
&0.05   & 0.12  & 0.01   & 0.02   & 4.80  & \underline{1.25}   & \underline{6.75}  & \underline{8.24}  & 13.31 & 14.95 & 7.55   & \underline{2.11}  & \underline{2.88}  & \underline{4.77}\\ 

\rowcolor[HTML]{E6F1FF}
\multicolumn{1}{l|}{\quad$\bullet$ Ours}                                &       This paper
&\textbf{0.11} & \textbf{0.22}  & \textbf{0.07} & \textbf{0.10} & \textbf{6.00}  & \textbf{2.00} & \textbf{6.89}  & \textbf{8.41}  & 13.58 & \textbf{15.94} & \textbf{7.94}  & \textbf{2.13}  & \textbf{2.92}  & \textbf{5.10}\\

\bottomrule[1pt]
\end{tabular}
}
}
\label{tab:tab1}
\vspace{-1.5mm}
\end{table*}

\begin{table*}[t]
\centering
\vspace{0mm}
\caption{Comparisons with state-of-the-art methods on the KITTI-C \emph{validation} set (severity level 5) with MonoGround as the  base model. Due to the space limit, the complete results of three categories are provided in Tab. \ref{tab:complete_monoground} of Appendix. \ref{sm:experiment}.}
\vspace{-1mm}
\resizebox{0.9\textwidth}{!}{
{
\fontsize{20}{25}\selectfont
\begin{tabular}{c|c|ccc|ccc|cccc|ccc|c}
\toprule[1pt]
\multicolumn{16}{c}{Car Category}\\
\toprule[1pt]
\multirow{2}{*}{\centering\textbf{Method}}  & \multirow{2}{*}{\centering\textbf{Reference}}   & \multicolumn{3}{c|}{Noise}              & \multicolumn{3}{c|}{Blur}                          & \multicolumn{4}{c|}{Weather}                       & \multicolumn{3}{c|}{Digital}                       &     \multirow{2}{*}{\centering\textbf{Avg}}       \\
 &  & Gauss.       & Shot       & Impul.     & Defoc.     & Glass      & Motion      & Snow       & Frost      & Fog        & Brit.      & Contr.    & Pixel       & Sat.         \\

\toprule[1pt]
\multicolumn{1}{c|}{MonoGround}                                &       CVPR'22
&0.00 & 0.00  & 0.00  & 0.00 & 11.63 & 0.29 & 1.95  & 6.59  & 3.14  & 19.25 & 0.00  & 4.66  & 3.74  & 3.94\\
\multicolumn{1}{l|}{\quad$\bullet$ TENT}                                &       ICLR'21
&6.82 & 14.81 & 8.21  & 4.88 & 28.38 & 2.65 & 23.92 & 28.08 & 33.06 & 36.70 & 20.22 & 30.63 & 33.27 & 20.90\\ 
\multicolumn{1}{l|}{\quad$\bullet$ EATA}                                &       ICML'22
&7.12 & 15.26 & 8.81  & 5.09 & 29.08 & 2.52 & 24.18 & 28.03 & 33.43 & 36.78 & 21.61 & 30.50 & 33.42 & 21.22\\ 
\multicolumn{1}{l|}{\quad$\bullet$ DeYO}                                &       ICLR'24
&7.35 & 15.72 & 9.38  & 5.74 & 30.01 & 2.99 & 25.03 & 28.55 & 34.32 & 37.31 & 23.41 & 30.99 & 34.16 & 21.92\\
\multicolumn{1}{l|}{\quad$\bullet$ MonoTTA}                                &       ECCV'24
&\underline{7.88} & \underline{16.73} & \underline{10.35} & \underline{5.97} & \underline{31.19} & \underline{3.06} & \underline{25.24} & \underline{28.99} & \underline{34.85} & \underline{37.82} & \underline{25.00} & \underline{31.61} & \underline{34.79} & \underline{22.57}\\ 

\rowcolor[HTML]{E6F1FF}
\multicolumn{1}{l|}{\quad$\bullet$ Ours}                                &       This paper
&\textbf{9.72} & \textbf{18.88} & \textbf{12.74} & \textbf{7.24} & \textbf{33.02} & \textbf{5.24} & \textbf{28.50} & \textbf{30.73} & \textbf{37.27} & \textbf{39.40} & \textbf{28.34} & \textbf{33.22} & \textbf{37.24} & \textbf{24.73}\\

\bottomrule[1pt]
\end{tabular}
}
}
\label{tab:tab2}
\vspace{-5.5mm}
\end{table*}

\vspace{-0.1cm}
\subsection{Semantic-Guided Normal Field Constraint}
\label{sec:5.2}
\vspace{-0.05cm}
Despite progress in uncertainty modeling, current methods struggle to handle geometric uncertainty in TTA scenarios due to critical limitations:
1) \textit{Model Collapse}: Direct optimization of model-predicted uncertainty leads to the degenerated predictor, as discussed in Sec. \ref{sec:observation}.
2) \textit{Substantial Overhead:} Geometry-aware methods typically require additional data or offline training for uncertainty estimation, limiting their real-time applicability \cite{poggi2020uncertainty,hornauer2022gradient}.
To overcome these challenges, we propose an efficient normal field constraint derived from a single image, which enhances the geometric coherence of 3D predictions, thereby reducing the uncertainty stemming from the unstable geometric representation, as shown in Fig. \ref{fig:pipeline}(c).

\noindent\textbf{Normal Field.} Given a depth map \( D \), we restore it to the original image resolution using bilinear interpolation, allowing for the alignment with the pixel grid. Then, we compute the spatial gradients of the depth map using efficient Sobel operators \cite{kanopoulos1988design}, which approximate the rate of change in the depth values along horizontal and vertical directions:
\begin{equation}
    \nabla D_x = \mathbf{S}_x \ast D, \quad \nabla D_y = \mathbf{S}_y \ast D,
\end{equation}
\noindent where \( \mathbf{S}_x \) and \( \mathbf{S}_y \) denote horizontal and vertical Sobel kernels, respectively. These gradient maps capture the variation in depth across neighboring pixels. The surface normal field, which encodes the orientation of the surface at each pixel, is then derived from the gradients as:
\begin{equation}
    \mathbf{N}(u,v) = \frac{1}{\sqrt{\nabla D_x^2 + \nabla D_y^2 + 1}} \begin{bmatrix} -\nabla D_x \\ -\nabla D_y \\ 1 \end{bmatrix},
\end{equation}
\mbox{where $\mathbf{N}(u,v)$ denotes the normal orientation at pixel $(u,v)$.}

\noindent\textbf{Normal Consistency Loss.} To quantify geometric uncertainty, we employ an edge-aware {\bf N}ormal {\bf C}onsistency {\bf L}oss (NCL), that encourages smoothness in the local surface by penalizing inconsistencies between neighboring pixels:
\begin{equation}
    \begin{aligned}
    \small
        \psi_x(u,v) = \|2\mathbf{N}(u,v) - \mathbf{N}(u+1,v) - \mathbf{N}(u-1,v)\|_2^2, \\ \small
    \psi_y(u,v)=\|2\mathbf{N}(u,v) - \mathbf{N}(u,v+1) - \mathbf{N}(u,v-1)\|_2^2,
    \end{aligned}
\end{equation}
where the smoothness terms $\psi_x(u,v), \psi_y(u,v)$ enforce horizontal and vertical consistency in the surface normal field. The total normal consistency loss is then given by: 
\begin{equation}
\small
    \mathcal{L}_\text{NCL}(u,v) = (\psi_x(u,v) + \psi_y(u,v)) \cdot \exp(-\|\nabla \text{I}(u,v)\|_2),
\label{eq:NC loss}
\end{equation}
where the edge-aware weighting term \( \exp(-\|\nabla \text{I}(u,v)\|_2) \) \( =\exp(-\sqrt{|\mathbf{S}_x \ast \text{I}(u,v)|^2+|\mathbf{S}_y \ast \text{I}(u,v)|^2})\) preserves discontinuities at boundaries while enforcing smoothness in homogeneous regions. NCL encourages the model to learn a spatially consistent normal field, thereby reducing uncertainty stemming from unstable 3D representations.

\noindent\textbf{Semantic Guidance.} To enforce geometric–semantic coherence, we generate masks by integrating 2D bounding boxes with semantic predictions, ensuring synchronized focus regions for dual-branch uncertainty optimization.
Let \( \{\mathcal{B}_i\}_{i=1}^n \) denote detected bounding boxes with scores \( \{s_i\}_{i=1}^n \) and semantic uncertainties $\{U_i \}_{i=1}^n$ derived from the Conjugate Focal Loss ($\mathcal{L}_\text{CFL}$ in Equ. \ref{eq:CF loss}). To ensure reliable supervision, we select boxes with low semantic uncertainty using an exponentially moving average threshold:

\vspace{-0.15cm}
\begin{equation}
R = \{i|U_i\leq \overline{U}\}, \;\overline{U} \leftarrow \beta \cdot\frac{\Sigma_{i=1}^{n} U_i}{n}+ (1-\beta)\cdot\overline{U},
\label{eq:ema}
\end{equation}
where $\beta \in [0, 1]$ is a moving average factor (set to 0.1 in default). We then construct the region mask as follows:
\begin{equation}
\mathcal{M}(u, v)=  \max_{i\in R} s_i\cdot\mathbb{I}_{\text {inside }}\left(u, v \mid \mathcal{B}_i\right),
\label{eq:mask}
\end{equation}
where \( \mathbb{I}_{\text{inside}}(u, v \mid \mathcal{B}_i) \) is an indicator function that returns 1 if pixel \( (u,v) \) is inside \( \mathcal{B}_i \), and 0 otherwise. This semantic-guided mask ensures that only low semantic-uncertainty regions contribute to the normal field constraint, enhancing the reliability of the geometric constraint.

\vspace{-0.1cm}
\subsection{Overall Objective}

The overall objective integrates the conjugate focal loss with the normal consistency constraint into a unified framework, simultaneously addressing semantic and geometric uncertainties. This dual optimization enables a complementary feedback loop: low-uncertainty spatial location enhances the model's ability to perform precise semantic classification, while confident semantic predictions in turn improve spatial understanding. The procedure is as follows:
\begin{equation}
    \min _{\theta} \, \Sigma_{x\in \text{I}} \mathcal{L}_\text{CFL}(x)+\lambda \Sigma_{(u,v)\in \text{I}}\mathcal{M}(u,v)\cdot\mathcal{L}_\text{NCL}(u,v), 
\label{eq:minimization}
\end{equation}
where $x$ iterates over all detected objects and $(u, v)$ spans all pixels in the image $\text{I}$. $\mathcal{L}_\text{CFL}$, $\mathcal{L}_\text{NCL}$, and $\mathcal{M}$ are defined in Equ. \ref{eq:CF loss}, Equ. \ref{eq:NC loss}, and Equ. \ref{eq:mask}. $\lambda=0.7$ by default.

%% file: sec/4_experiment.tex
\section{Experiments}
\subsection{Experimental Setup}
For a fair comparison, we follow the identical evaluation pipeline with prior work \cite{lin2025monotta}, including baseline models, training recipes, and evaluation protocols.

\noindent\textbf{Datasets.} We conduct experiments on the KITTI~\cite{geiger2012we} and nuScenes~\cite{caesar2020nuscenes} datasets. For KITTI, we use the KITTI-C version, which includes 13 distinct data corruption types~\cite{hendrycks2018benchmarking} and five severity levels per type. 
Results represent the average performance across three difficulty levels, \ie Easy, Moderate, and Hard.
Note that we also provide more results of different severity levels of KITTI in Appendix. \ref{sm:levels}. For nuScenes, we adopt the front-view images and construct the Daytime, Night, Sunny, and Rainy scenarios via their scene descriptions following~\cite{liu2023bevfusion}. Based on these scenes, we define 4 real-world adaptation tasks, \ie,~Daytime $\leftrightarrow$ Night and Sunny $\leftrightarrow$ Rainy. Following the MonoTTA, we transfer the nuScenes dataset into the KITTI format. More details are provided in Appendix. \ref{sm:data}.

\noindent\textbf{Compared Methods.} Based on two representative base models MonoFlex ~\cite{zhang2021objects} and MonoGround \cite{qin2022monoground}, we compare our DUO with several state-of-the-art methods: EATA~\cite{eata} identifies reliable samples during entropy training. DeYO \cite{deyo} leverages probability variations under augmentations as an additional cue to enhance entropy optimization. MonoTTA \cite{lin2025monotta} boosts probabilities for high-score classes while applying negative learning to low-score classes.

\noindent\textbf{Implementation Details.} We implement our method and other baselines in PyTorch~\cite{paszke2019pytorch}. We employ the Stochastic Gradient Descent (SGD) optimizer with the same learning rate as MonoTTA, a momentum of 0.9, and a batch size of 16 for KITTI, 4 for nuScenes. Parameters $\lambda$, $\alpha$, $\gamma$ are assigned default values of 0.7, 4, and 2, respectively.

\noindent\textbf{Evaluation Protocols.}
We report experimental results using the Average Precision for 3D bounding boxes, denoted as  $\text{AP}_{3D|R_{40}}$ \cite{simonelli2019disentangling}. Results represent the mean values across three difficulty levels, with Intersection over Union thresholds set to 0.5 for Cars and 0.25 for both Ped. and Cyc.

\begin{table}[!t]
\begin{center}
\vspace{-0.3cm}
\caption{\label{tab:nus_day} Comparison with baselines on \textbf{D}aytime $\leftrightarrow$ \textbf{N}ight and \textbf{S}unny $\leftrightarrow$ \textbf{R}ainy of {nuScenes} dataset, regarding ${\rm AP}_{{\rm 3D}|{\rm R}_{40}}$.}
\vspace{-0.35cm}
\resizebox{\linewidth}{!}{
         \begin{tabular}{l|ccc|cccccccc}
         \toprule
         \multirow{2}{*}{\centering\textbf{Method}} & 
         \multicolumn{3}{c|}{MonoFlex} &
         \multicolumn{3}{c}{MonoGround}
         \\
         \cmidrule(lr){2-4} \cmidrule(lr){5-7} 
         & D $\rightarrow$ N & N $\rightarrow$ D & Avg. & D $\rightarrow$ N & N $\rightarrow$ D & Avg. \\
         \midrule
         Source model & 1.53          & 2.75          & 2.14          & 6.97           & 1.09          & 4.03  \\ ~~$\bullet~$ TENT
         & 3.33          & 3.45          & 3.39          & 8.36           & 1.66          & 5.01 \\
         ~~$\bullet~$ DeYO  & 4.72          & 4.87          & 4.79          & 11.01          & 1.40          & 6.21   \\
         ~~$\bullet~$ MonoTTA  & 6.92          & 3.68          & 5.30          & 13.61          & 1.29          & 7.45   \\
         \rowcolor[HTML]{E6F1FF}
         ~~$\bullet~$ Ours & \textbf{9.05} & \textbf{5.41} & \textbf{7.23} & \textbf{15.70} & \textbf{1.91} & \textbf{8.81}  \\
         \midrule 
         \centering\textbf{Method} &  S $\rightarrow$ R & R $\rightarrow$ S & Avg. & S $\rightarrow$ R & R $\rightarrow$ S & Avg. \\
         \midrule
         Source model & 6.86           & 10.91          & 8.89           & 8.03           & 7.44          & 7.73  \\
         ~~$\bullet~$ TENT & 8.53           & 11.61          & 10.07          & 9.94           & 8.71          & 9.33 \\
         ~~$\bullet~$ DeYO & 9.33           & 12.04          & 10.68          & 10.54          & 9.10          & 9.82  \\
         ~~$\bullet~$ MonoTTA  & 9.47           & 12.55          & 11.01          & 10.89          & 8.90          & 9.90 \\
         \rowcolor[HTML]{E6F1FF}
         ~~$\bullet~$ Ours  &  \textbf{11.54} & \textbf{13.21} & \textbf{12.38} & \textbf{12.88} & \textbf{9.54} & \textbf{11.21}  \\
         \bottomrule
\end{tabular}}
\end{center}
\vspace{-0.3cm}
\end{table}

\subsection{Main Results}
\noindent\textbf{Evaluation on Corruption Shifts.}
We first compare our DUO with previous methods on the KITTI-C dataset at severity level 5. Due to space constraints, detailed comparisons with other severity levels are provided in the Appendix. \ref{sm:experiment}. The results, reported in Tab.~\ref{tab:tab1} \&~\ref{tab:tab2}, reveal several observations: 1) Under test-time distribution shifts, pre-trained detectors experience significant performance degradation across all categories. 2) Existing TTA methods partially mitigate the adverse effects of distribution shifts in M3OD, but their performance remains suboptimal as they cannot address the dual uncertainty. 3) Our method consistently outperforms compared approaches across all categories and base models. Notably, the proposed DUO achieves the best or comparable performance under 13 types of corruptions, with performance gains of $\textbf{+2.1}$ and $\textbf{+2.2}$ ${\rm AP}_{{\rm 3D}|{\rm R}_{40}}$ in the Car category for two different models. These observations validate the crucial role of our adaptation framework in addressing dual uncertainty in M3OD models and enhancing robustness against distribution shifts.

\noindent\textbf{Evaluation on Real-World Scenario.} We further evaluate different methods on four real-world scenarios, as shown in Tab.~\ref{tab:nus_day}. The experimental results yield the following observations: 
1) Under real-world shifts, the pre-trained model still suffers severe performance degradation.
2) For Sunny $\leftrightarrow$ Rainy adaptation tasks on MonoFlex, existing methods achieve only marginal improvements, whereas our method boosts performance by $\textbf{+4.7}$ ${\rm AP}_{{\rm 3D}|{\rm R}_{40}}$.
3) Our DUO brings significant performance improvement on two base models, maintaining the best performance across all four scenarios, further demonstrating its effectiveness and superiority.

%% file: sec/5_ablation.tex
\subsection{Ablation Study}
\label{sec:ablation}
In this section, without loss of generality, we conduct ablation studies on Monoflex under Gaussian shift in KITTI-C for the sake of brevity. Focusing on two pivotal components of DUO, \textit{Conjugate Focal Loss} and \textit{Normal Field Constraint}, we perform extensive experiments to analyze their independent roles and complementary effects, gaining insights into key factors contributing to their effectiveness.
\begin{table}[!t]
\centering
\vspace{-0.3cm}
\caption{Effects of components in our method.  We conduct ablation studies of the conjugate focal loss $\mathcal{L}_\text{CFL}$, normal consistency loss $\mathcal{L}_\text{NCL}$, and semantic guidance $\mathcal{M}$ on the {KITTI-C validation} set. ''Src." denotes the source model without adaptation.}
\vspace{-0.1cm}
\resizebox{\linewidth}{!}{
\begin{tabular}{cccc|cccc|ccccccc}
         \toprule[1pt]
         \multirow{2}{*}{Src.} & \multirow{2}{*}{$\mathcal{L}_\text{CFL}$} & \multirow{2}{*}{$\mathcal{L}_\text{NCL}$} &  \multirow{2}{*}{$\mathcal{M}$} &
         \multicolumn{4}{c|}{MonoFlex} &
         \multicolumn{4}{c}{MonoGround}
         \\
         \cmidrule(lr){5-8} \cmidrule(lr){9-12} 
         & & & & Car & Pedes. & Cyc. & Avg. & Car & Pedes. & Cyc. & Avg. \\
         \midrule
          \ding{52} &  &  &  & 4.54  & 0.88  & 0.83  & 2.08 & 3.94  & 1.79  &0.52  & 2.08    \\
          & \ding{52} &  &  & 20.98  & 6.60  & 4.32  & 10.63  & 22.89  & 8.86  & 2.51  & 11.42   \\
          & & \ding{52} & & 12.38  & 4.63  & 2.82  & 6.61  & 15.69  & 6.40  & 1.91  & 8.00   \\
          & \ding{52} & \ding{52} & & \underline{22.17}  & \underline{6.99}  & 4.72  & \underline{11.29}  & \underline{24.29}  & \underline{9.27}  & 2.78  & \underline{12.11}   \\
         & & \ding{52} & \ding{52} & 16.49  & 6.23  & \underline{4.87}  & 9.20  & 19.68  & 7.98  & \underline{2.94}  & 10.20   \\
         \rowcolor[HTML]{E6F1FF}
         & \ding{52} & \ding{52} & \ding{52} & \textbf{22.97}  & \textbf{7.19}  & \textbf{5.10}  & \textbf{11.75}  & \textbf{24.73}  & \textbf{9.62}  & \textbf{3.02}  & \textbf{12.46}   \\
         \bottomrule[1pt]
         \end{tabular}
         }
\label{tab:component}
\end{table}

\begin{figure}[t]
\setlength{\abovecaptionskip}{0cm} 
\begin{center}
\includegraphics[width=1.0\linewidth]{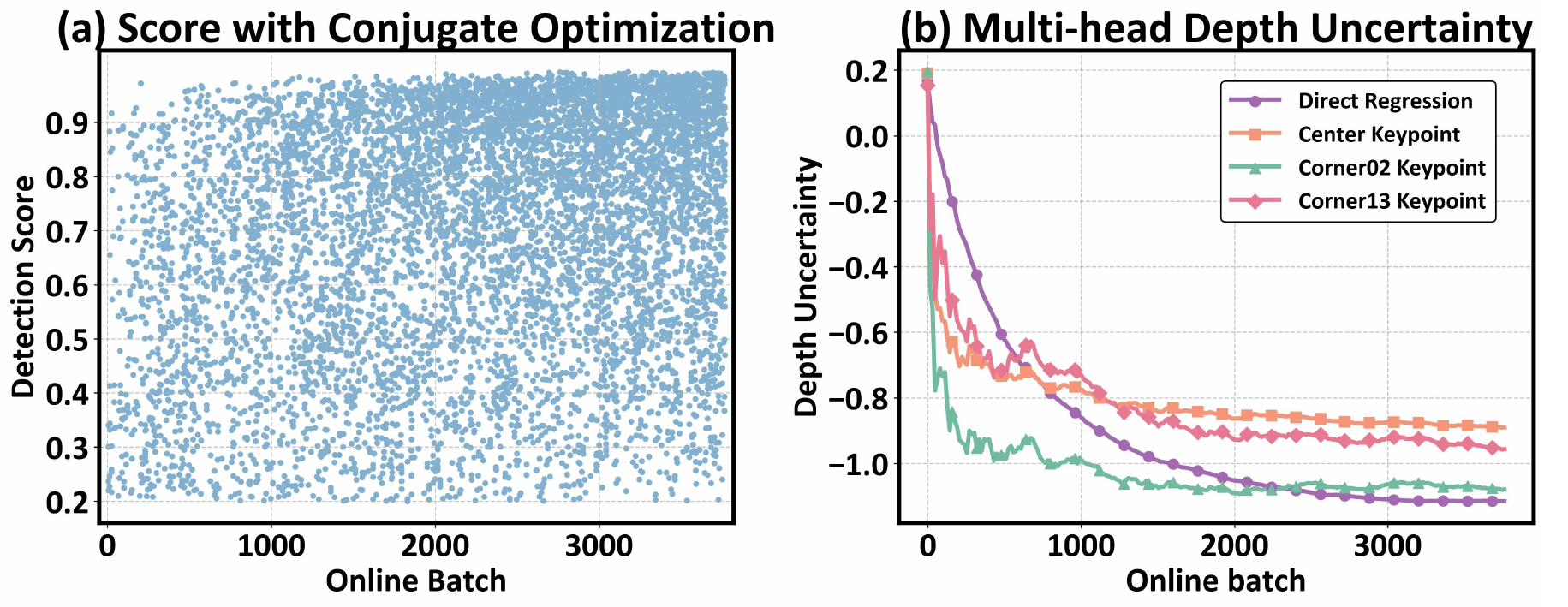}
\end{center}
\vspace{0cm}
\caption{Effects of components in our method. ({\it a}) shows the distribution of detection scores during adaptation with conjugate focal loss. ({\it b}) records the variation of depth uncertainty in different heads during adaptation with normal field constraints.}
\vspace{-0.3cm}
\label{fig:ablation}
\end{figure}

\begin{figure}[!t]
\setlength{\abovecaptionskip}{0cm} 
\begin{center}
\includegraphics[width=1.0\linewidth]{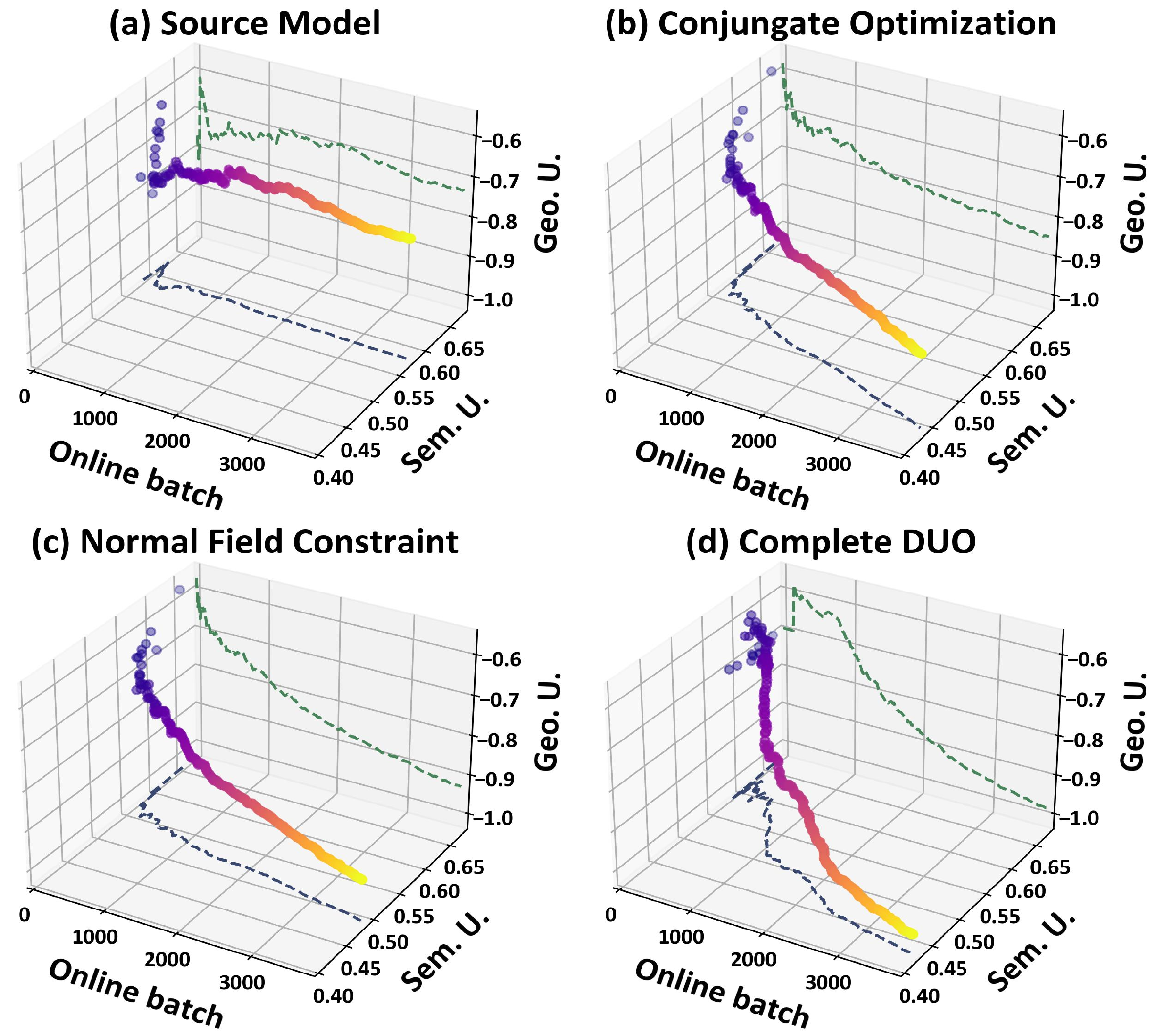}
\end{center}
\vspace{-0.1cm}
\caption{Impacts of different components on dual uncertainty. Colored lines in three-dimensional space show the cumulative average of dual uncertainty during the adaptation process. Green and blue lines in two-dimensional planes record geometric uncertainty and semantic uncertainty, respectively.}
\vspace{-0.4cm}
\label{fig:complementary}
\end{figure}

\noindent {\bf Effectiveness of Components.}
We investigate the impact of individual components by comparing the full method with variations that omit key parts. 
As shown in Tab. \ref{tab:component}, incorporating either the conjugate focal loss or the normal field constraint significantly enhances detection performance, yielding average gains of $+8.9$ and $+7.6$ ${\rm AP}_{{\rm 3D}|{\rm R}_{40}}$, respectively. Notably, using the normal consistency loss ($\mathcal{L}_\text{NCL}$) alone results in unstable, marginal improvements; its effectiveness depends on being combined with $\mathcal{M}$, highlighting the necessity of semantic guidance for effective geometric uncertainty reduction. These findings underscore the effectiveness of each component.

Furthermore, a comparison of score distributions in Fig. \ref{fig:ablation}(a) and Fig. \ref{fig:observe}(c) shows that the conjugate focal loss consistently boosts detection scores, ensuring effective training for challenging low-score objects. Similarly, Fig. \ref{fig:ablation}(b) versus Fig. \ref{fig:observe}(d) demonstrates that our normal field constraint consistently reduces geometric uncertainty across all heads, preventing the model collapse observed in baseline methods. Therefore, these innovations effectively overcome the limitations of existing methods identified in Sec. \ref{sec:observation}, leading to more robust and reliable detection.

\noindent
{\bf Complementary Effects.}
As shown in Tab. \ref{fig:ablation}, our complete method, which integrates all components, consistently achieves the best detection performance, demonstrating the compatibility of different components. To further investigate the complementary effects of our dual-branch design, we visualize the dual-uncertainty optimization process. Specifically, comparing Fig. \ref{fig:complementary}(a)\&(b), we observe that applying the conjugate focal loss significantly reduces semantic uncertainty, while interestingly, geometric uncertainty also shows a modest decline. A similar phenomenon is evident when employing the normal field constraint in Fig. \ref{fig:complementary}(c). This synchronous behavior suggests an intrinsic interdependence between the two types of uncertainty, validating the potential of joint optimization.

Moreover, by combining two innovations in our DUO framework, it achieves the fastest and most pronounced decrease in two uncertainties compared to individual components, as shown in Fig. \ref{fig:complementary}(d). These observations further validate that our dual-branch architecture creates a complementary loop for dual-uncertainty optimization, effectively harnessing their complementary effects.

\noindent\textbf{Robustness and Efficiency.} To validate the robustness and efficiency of our method, we extend our analysis by examining the sensitivity of key hyper-parameters and comparing running times in Appendix. \ref{sm:ablation}.

\subsection{Qualitative  Results}
\begin{figure}[t]
\setlength{\abovecaptionskip}{0cm} 
\begin{center}
\includegraphics[width=1.0\linewidth]{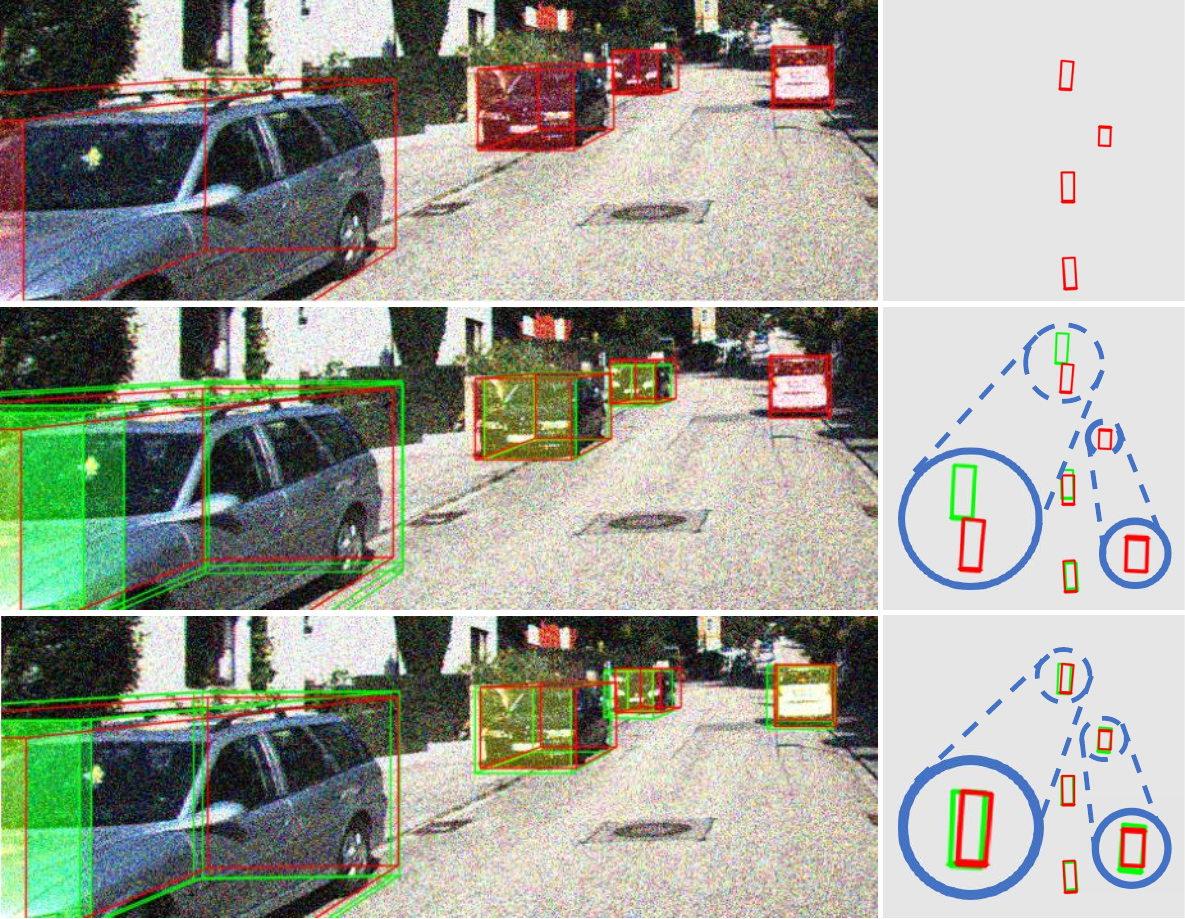}
\end{center}
\vspace{-0.1cm}
\caption{Qualitative examples on KITTI-C. In each row, we provide the front view ({\it left}) and the bird’s-eye view ({\it right}) visualizations. \textcolor{red}{Red} represents the ground truth of boxes, while \textcolor{green!80!black}{Green} represents the predictions. We circle some objects to highlight differences in predictions. The first row corresponds to the source model, the second to MonoTTA, and the third to our method.}
\vspace{-0.3cm}
\label{fig:quan}
\end{figure}

Based on the qualitative results shown in Fig. \ref{fig:quan}, our DUO framework produces predictions that are more precisely aligned with the ground-truth annotations. Compared with the latest SOTA method, DUO not only significantly improves the accuracy of 3D location estimation but also reduces missed detections for challenging distant or small-scale objects, which are typically prone to large geometric errors. These visualizations further confirm that our dual uncertainty optimization effectively adapts the source model, achieving precise and reliable 3D perception that closely matches the ground truth.

%% file: sec/6_conclusion.tex
\section{Conclusion and Future Work}
In this paper, we propose a synergistic dual-branch framework designed to address semantic-geometric uncertainty inherent in 3D vision systems under test-time domain shifts. Our approach derives a novel conjugate loss to offer an adaptive, label-free weighting mechanism for balanced training on semantic uncertainty. Meanwhile, it incorporates a normal consistency constraint to reduce uncertainty from inconsistent geometric representation. Our extensive experiments demonstrate that the proposed dual-branch optimization creates a complementary loop, consistently improving performance across diverse domain shifts. In future work, we tend to expand our framework beyond M3OD to cover a broader range of 3D vision tasks. We hope that our work will deepen the understanding of uncertainty-aware model adaptation while providing transferable insights for related fields, such as unsupervised learning.

\section*{Acknowledgements}
This work was supported by the Program of Beijing Municipal Science and Technology Commission
Foundation (No.Z241100003524010), in part by the National Natural Science Foundation of China
under Grant 62088102, in part by
AI Joint Lab of Future Urban Infrastructure sponsored by Fuzhou Chengtou New Infrastructure Group and Boyun Vision Co. Ltd, and in part by the PKU-NTU Joint Research Institute (JRI) sponsored by a donation from the Ng Teng Fong Charitable Foundation.

%% file: sec/7_appendix.tex
\appendix
\onecolumn
\begin{center}
{\LARGE \textbf{Adaptive Dual Uncertainty Optimization: Boosting Monocular 3D Object Detection under Test-Time Shifts\\ $\;$ \\ ————Appendix————}}
\end{center}
 The structure of Appendix is as follows:
\begin{itemize}
    \item Appendix~\ref{sm:proof} contains all missing proofs in the main manuscript.
    \item Appendix~\ref{sm:experiment} presents further experimental results on more corruption levels.
    \item Appendix~\ref{sm:ablation} provides additional ablation studies to validate the robustness and efficiency of our method.
    \item Appendix~\ref{sm:implementation} details the compared methods, model architecture, and datasets used for comparison.
\end{itemize}

\section{Theoretical Proof}
\label{sm:proof}
Below, we provide detailed proofs of the theoretical results presented in Sec. 5.1 of the main paper.

\noindent \textbf{Notation.} First, we recall the notation that we used in the main paper as well as this appendix: $x$ denotes an inputting test image and $y$ denotes the one-hot coding of the ground-truth label. $h_\theta$ denotes the model with its parameter set $\theta$ and $h\triangleq h_\theta (x)$. $s \triangleq e^{h_1}+...+e^{h_\text{c}}$ is the sum over the exponential outputs of the model and $\text{c}$ is the number of classes. $p\triangleq \texttt{softmax}(h)$ is the normalized probability over the classes. $\text{diag}(\cdot)$ denotes the diagonal matrix and $I$ denotes the identity matrix. We define the following two functions: $f(h)=\alpha\log s$, $g(h)= \alpha h+\alpha((1-p)^\gamma -1) \log p$.

\subsection{Legendre-Fenchel Structure} 
In this subsection, we demonstrate the equivalence of the vanilla focal loss with its Legendre-Fenchel structure.

\begin{equation}
\begin{aligned}
\mathcal{L}_{\text{FL}}(x, y) & =-\alpha(1-p)^\gamma \cdot y \log p \\
& =-\alpha(1-p)^\gamma \cdot y\left(\log \left(\begin{array}{c}
e^{h_1} \\
\vdots \\
e^{h_c}
\end{array}\right)-\log \left(e^{h_1}+\cdots+e^{h_c}\right)\right) \\
& =-\alpha(1-p)^\gamma \cdot y\cdot\log\left(\begin{array}{c}
e^{h_1} \\
\vdots \\
e^{h_c}
\end{array}\right) +\alpha(1-p)^\sigma y \log s \\
& =-\alpha(1-p)^\gamma \cdot y \cdot h+\alpha \cdot y \log s+\alpha\left((1-p)^\sigma-1\right) y \log s.
\end{aligned}
\end{equation}
Duo to $y$ is the one-hot vector, we have $y\log s = \log s$ and we further derive:
\begin{equation}
\begin{aligned}
\mathcal{L}_{\text{FL}}(x, y) & =\alpha \log s+\alpha y\left(-(1-p)^\gamma\cdot h + (1-p)^\gamma \log s-\log s\right) \\
& =\alpha \log s+\alpha y\left(-(1-p)^\gamma \cdot(h-\log s)-\log s\right) \\
& =\alpha \log s+\alpha y\left(-\log s-(1-p)^\gamma \log p\right) \\
& =\alpha \log s-y^{\top}\left(\alpha \log s+\alpha(1-p)^\gamma \log p\right) \\
& =\underbrace{\alpha \log s}_{f(h)}-y^{\top} \underbrace{\left(\alpha h+\alpha\left((1-p)^\gamma-1\right) \log p\right)}_{g(h)}.
\end{aligned}
\end{equation}
Therefore, the focal loss can be formulated as a classical convex conjugate structure, allowing for further analysis in the conjugate optimization framework. 

\subsection{Problem Reconstruction.} 
In this section, we demonstrate the invertibility of function $g$ to ensure the existence of a conjugate function and further reformulate the optimization problem into conjugate relationships. by the inverse function theorem, the local invertibility of $g$ is guaranteed if its Jacobian is non-singular. For simplicity, we demonstrate the positive definiteness of the Jacobian under the default setting $\gamma = 2$:

\noindent\textbf{Step 1: Jacobian of $g$.} 
Taking the gradient of $g$ with respect to $h$, we obtain:
\begin{equation}
\nabla_h g=I+\text{diag}(p-2-2(1-p)\log p)\cdot(\text{diag}(p)-pp^\top):= I + D\cdot H. 
\end{equation}
where $D=\text{diag}(p-2-2(1-p)\log p)$ and $H=\text{diag}(p)-pp^\top$.

\noindent\textbf{Step 2: Positive Semidefiniteness of $H$.} 
For any vector $\mathbf{v} \in \mathbb{R}^C$, the quadratic form of $\mathbf{H}$ is:
\begin{equation}
    \mathbf{v}^{\top} \mathbf{H} \mathbf{v}=\sum_{i=1}^C p_i v_i^2-\left(\sum_{i=1}^C p_i v_i\right)^2
\end{equation}
Let $V$ be a random variable that takes the value $v_i$ with probability $p_i$. The quadratic form simplifies to:
\begin{equation}
    \mathbf{v}^{\top} \mathbf{H} \mathbf{v}=\mathbb{E}\left[V^2\right]-(\mathbb{E}[V])^2=\operatorname{Var}(V)
\end{equation}
Since variance is always non-negative, we have:
\begin{equation}
    \mathbf{v}^{\top} \mathbf{H} \mathbf{v} \geq 0 \quad \forall \mathbf{v} \in \mathbb{R}^C
\end{equation}
Thus, $\mathbf{H}$ is positive semi-definite. Furthermore, using Gershgorin’s circle theorem, all eigenvalues of $H$ satisfy:
\begin{equation}
0 \leq \lambda(H) \leq \max _i\left\{2 p_i\left(1-p_i\right)\right\} \leq \frac{1}{2},
\end{equation}
where $2p_i\left(1-p_i\right)$ attains its maximum $\frac{1}{2}$ when $p_i=\frac{1}{2}$.

\noindent\textbf{Step 3: Bounding the Eigenvalues of $\nabla_h g$.} Since $D$ is diagonal, the matrix $D\cdot H$ can be seen as a row-scaled version of $H$. And the element of $H$ follows:
\begin{equation}
D_{i i}=p_i-2-2\left(1-p_i\right) \log p_i .
\end{equation}
Numerical analysis of the function $\phi(p_i)=p_i-2-2\left(1-p_i\right) \log p_i $ shows $D_{i i} \textgreater -2$ for $p_i\in \left(0,1\right)$. Therefore, the eigenvalues of $\nabla_h g$ satisfy:
\begin{equation}
\lambda\left(\nabla_h g\right) \geq 1+\lambda_{\min }(D) \cdot \lambda_{\min }(H) .
\end{equation}
Since $\lambda_{\min }(H) \geq 0$ and $D_{i i} \textgreater -2$, we can obtain a tighter bound:
\begin{equation}
\lambda\left(\nabla_h g\right) \geq 1-2 \cdot \lambda(H) \geq 1-2 \cdot \frac{1}{2}=0 .
\end{equation}

\noindent\textbf{Step 4: Conclusion.} As all eigenvalues of $\nabla_h g$ are non-negative, the Jacobian is positive semidefinite. In particular, as long as $p$ is not degenerate (i.e., no prediction has probability exactly 0 or 1), $H$ is full rank on its subspace, ensuring that $\nabla_h g$ is non-singular in a neighborhood of $h$. Thus, by the inverse function theorem, $g$ is locally invertible.

This invertibility guarantees the existence of a conjugate function $f^{*}$, which equals to the minimization value of the objective:
\begin{equation}
    \min_h \{f(h)-y^\top g(h)\} = \min_{z=g(h)} \{f\circ g^{-1}(z)-y^\top z\}=f^{*}(y).
\end{equation}
Under the common assumption that the representation $h$ pre-trained from the large source dataset is already close to a local optimal solution $h_0$, we can convert the problem into the following conjugate relationships:
\begin{equation} 
 \begin{array}{l} f \circ g^{-1}(z) - y^{\top} z = f^*(y), \ \nabla_z (f \circ g^{-1}) = y. \end{array}  
 \label{eq:conjugate}
 \end{equation}
\subsection{Conjugate Focal Loss} 
In this subsection, we need to derive the estimation of $y$ from the conjugate conditions in Equ. \ref{eq:conjugate}. For the derivative of $g$, we differentiate the two summation components separately:
\begin{equation}
    g(h)=\alpha\,h+\alpha\,\phi(p),\quad\text{with}\quad \phi(p)=((1-p)^\gamma-1)\cdot\log p.
\end{equation}
For the first part, we have $\nabla_h h=I$. For the second part, we utilize the chain rule to derive the following format:
\begin{equation}
    \nabla_h\phi=\nabla_p\phi\cdot\nabla_h p.
\end{equation}
We calculate the $\nabla_p\phi$ as follows:
\begin{equation}
    \nabla_p\Bigl(((1-p)^\gamma-1)\log p\Bigr)
=\frac{(1-p)^\gamma-1}{p}-\gamma\,(1-p)^{\gamma-1}\log p.
\end{equation}
We calculate the $\nabla_h p$ as follows:
\begin{equation}
    \nabla_h p=\text{diag}(p)-pp^\top.
\end{equation}
Thus, the Jacobian of \(g\) is given by
\begin{equation}
    \nabla_h g(h)=\alpha\Bigl[I+(\frac{(1-p)^\gamma-1}{p}-\gamma(1-p)^{\gamma-1}\log p)\cdot(\text{diag}(p)-pp^\top)\Bigr].
\end{equation}
For the composite function \(f\circ g^{-1}\), the chain rule gives
\begin{equation}
    \nabla_z(f\circ g^{-1})(z)=\nabla_h f(h)\,\bigl(\nabla_h g(h)\bigr)^{-1}.
\end{equation}
Evaluating at \(h_0\) (with \(p=\operatorname{softmax}(h_0)\)) leads to
\begin{equation}
    y_0=\bigl(\nabla_h g(h_0)\bigr)^{-1}\nabla_h f(h_0)
=\Bigl[I+(\frac{(1-p)^\gamma-1}{p}-\gamma(1-p)^{\gamma-1}\log p)\cdot(\text{diag}(p)-pp^\top)\Bigr]^{-1}p.
\end{equation}
And we utilize the Taylor's Formula to approximate the value (neglecting higher-order terms of $p$):
\begin{equation}
    (1-p)^\gamma \approx 1-\gamma p+\frac{\gamma(\gamma-1)}{2}\text{diag}(pp^T), (1-p)^{\gamma-1}\log p \approx (1-(\gamma-1)p)\log p.
\end{equation}
After algebraic manipulation (and neglecting higher-order terms in \(p \) and \( \log p\)), we obtain
\begin{equation}
    y_0\approx \Bigl(I+\gamma\Bigl((1-\log p)\,pp^\top-\log p\;\operatorname{diag}(p)\Bigr)\Bigr)^{-1}p.
\end{equation}
Thus, by applying the chain rule and approximating the Jacobian of \(g\) to higher order, we obtain
\begin{equation}
    y_0\triangleq\left.\frac{\nabla_h (f\circ g^{-1})}{\nabla_h z}\right|_{z=g(h_0)}
\approx \Bigl(I+\gamma(1-\log p)\,pp^\top-\gamma\log p\;\operatorname{diag}(p)\Bigr)^{-1}p.
\end{equation}
Finally, we substitute this estimation into Equ. \ref{eq:conjugate}, yielding Conjugate Focal Loss:
\begin{equation}
\mathcal{L}_{\text{CFL}}(x)=f(h)-y_0^{\top} g(h) 
=  -\alpha (1-p)^\gamma (I+\gamma(1-\log p)\cdot p^{\top} p 
 -\gamma \log p \cdot \text{diag}(p))^{-1} p \log p.
\end{equation}

\begin{table*}[b]
\centering
\caption{Comparisons with state-of-the-art methods on the KITTI-C \emph{validation} set \textbf{(severity level 3)} in Car category. We highlight the best and second results with {\bf bold} and \underline{underline} respectively.}
\resizebox{\textwidth}{!}{
{
\fontsize{20}{25}\selectfont
\begin{tabular}{c|c|ccc|ccc|cccc|ccc|c}
\toprule[1pt]
\multirow{2}{*}{\centering\textbf{Method}}  & \multirow{2}{*}{\centering\textbf{Reference}}   & \multicolumn{3}{c|}{Noise}              & \multicolumn{3}{c|}{Blur}                          & \multicolumn{4}{c|}{Weather}                       & \multicolumn{3}{c|}{Digital}                       &     \multirow{2}{*}{\centering\textbf{Avg}}       \\
 &  & Gauss.       & Shot       & Impul.     & Defoc.     & Glass      & Motion      & Snow       & Frost      & Fog        & Brit.      & Contr.    & Pixel       & Sat.         \\

\toprule[1pt]
\multicolumn{1}{c|}{MonoFlex}                                &       CVPR'21
&0.63 & 0.49 & 0.63 & 1.09 & 26.10 & 0.71 & 14.21 & 15.88 & 10.16 & 27.88 & 4.41 & 11.61 & 39.25 & 11.77\\
\multicolumn{1}{l|}{\quad$\bullet$ TENT}                                &       ICLR'21
&17.99 & 26.99 & 22.29 & 13.46 & 35.73 & 9.36 & 32.52 & 30.99 & 38.13 & 40.67 & 39.28 & 34.46 & 43.37 & 29.63\\ 
\multicolumn{1}{l|}{\quad$\bullet$ EATA}                                &       ICML'22
&18.21 & 27.52 & 22.83 & 14.86 & 36.01 & 13.98 & 33.11 & 31.45 & 38.35 & 40.62 & 39.55 & 35.23 & 43.44 & 30.39\\ 
\multicolumn{1}{l|}{\quad$\bullet$ DeYO}                                &       ICLR'24
&18.36 & \underline{28.49} & 23.15 & 15.04 & \underline{36.44} & 16.38 & 33.67 & 31.32 & 38.57 & 40.75 & 39.93 & 35.81 & \underline{43.58} & 30.89\\
\multicolumn{1}{l|}{\quad$\bullet$ MonoTTA}                                &       ECCV'24
&\underline{19.64} & 28.37 & \underline{24.45} & \underline{17.79} & 35.91 & \underline{17.20} & \underline{34.11} & \underline{31.78} & \underline{39.45} & \underline{40.83} & \underline{40.74} & \underline{36.27} & 43.46 & \underline{31.54}\\ 

\rowcolor[HTML]{E6F1FF}
\multicolumn{1}{l|}{\quad$\bullet$ Ours}                                &       This paper
&\textbf{21.18} & \textbf{29.43} & \textbf{25.43} & \textbf{19.09} & \textbf{36.85} & \textbf{18.88} & \textbf{35.32} & \textbf{31.96} & \textbf{39.77} & \textbf{41.64} & \textbf{41.71} & \textbf{36.73} & \textbf{43.93} & \textbf{32.46}\\
\toprule[1pt]
\multicolumn{1}{c|}{MonoGround}                                &       CVPR'22
&0.51 & 0.52 & 0.86 & 2.47 & 25.71 & 0.35 & 10.68 & 9.99 & 5.59 & 32.31 & 0.81 & 14.94 & 36.06 & 10.83\\
\multicolumn{1}{l|}{\quad$\bullet$ TENT}                                &       ICLR'21
&20.01 & 31.16 & 25.56 & 17.72 & 38.63 & 10.47 & 33.58 & 30.83 & 38.06 & 42.78 & 40.11 & 39.56 & \textbf{45.49} & 31.84\\ 
\multicolumn{1}{l|}{\quad$\bullet$ EATA}                                &       ICML'22
&20.36 & 31.84 & 26.63 & 18.39 & 38.77 & 14.21 & 34.03 & 31.08 & 37.93 & 42.32 & 40.32 & 39.57 & 45.30 & 32.37\\ 
\multicolumn{1}{l|}{\quad$\bullet$ DeYO}                                &       ICLR'24
&20.80 & 32.31 & 27.32 & 19.33 & 38.63 & 15.37 & \underline{34.58} & 31.43 & 33.95 & \textbf{42.97} & 40.33 & 39.83 & 45.20 & 32.47\\
\multicolumn{1}{l|}{\quad$\bullet$ MonoTTA}                                &       ECCV'24
&\underline{22.10} & \underline{33.93} & \underline{28.35} & \underline{22.49} & \underline{39.88} & \underline{16.64} & 32.71 & \underline{32.42} & \underline{39.93} & 42.69 & \underline{40.61} & \underline{39.92} & 44.85 & \underline{33.58}\\ 

\rowcolor[HTML]{E6F1FF}
\multicolumn{1}{l|}{\quad$\bullet$ Ours}                                &       This paperX
&\textbf{23.65} & \textbf{34.79} & \textbf{29.23} & \textbf{23.08} & \textbf{40.63} & \textbf{18.66} & \textbf{34.75} & \textbf{33.38} & \textbf{40.39} & \textbf{42.97} & \textbf{41.64} & \textbf{40.53} & \underline{44.91} & \textbf{34.51}\\

\bottomrule[1pt]
\end{tabular}
}
}
\label{tab:level3}
\end{table*}

\section{Further Experiments}
\label{sm:experiment}
In this section, we broaden our investigation by evaluating our method across a variety of shift severity levels. To this end, we conduct experiments on corruption scenarios with shift level 1, 3, 5. These experiments allow us to thoroughly examine the robustness of our approach under different severity levels of distribution shifts.
\subsection{Different Severity Level Corruption}
\label{sm:levels}
We further provide more discussions surrounding high-severity data corruptions (\ie 3 and 1) based on the experimental results shown in Tab.~\ref{tab:level3}\&\ref{tab:level1}, which clearly gives additional observations: 
1) With the escalation of severity level, the source models suffer a larger performance decline within various corruptions. For instance, the pre-trained models of MonoFlex and MonoGround achieve obvious performance drop  from level 1 to level 3, which significantly heightens the challenge for test-time adaptation.
2) Existing TTA methods struggle to recover performance under such extreme corruptions, highlighting the limitations of conventional uncertainty optimization approaches.
3) Despite these challenges, our DUO framework consistently achieves the best average performance across all corruption types. This robust performance demonstrates that our dual uncertainty optimization framework effectively stabilizes both the semantic classification and spatial perception branches, providing reliable adaptation even under different shift levels.
\begin{table*}[t]
\centering
\caption{Comparisons with state-of-the-art methods on the KITTI-C \emph{validation} set \textbf{(severity level 1)} in Car category. We highlight the best and second results with {\bf bold} and \underline{underline} respectively.}
\resizebox{\textwidth}{!}{
{
\fontsize{20}{25}\selectfont
\begin{tabular}{c|c|ccc|ccc|cccc|ccc|c}
\toprule[1pt]
\multirow{2}{*}{\centering\textbf{Method}}  & \multirow{2}{*}{\centering\textbf{Reference}}   & \multicolumn{3}{c|}{Noise}              & \multicolumn{3}{c|}{Blur}                          & \multicolumn{4}{c|}{Weather}                       & \multicolumn{3}{c|}{Digital}                       &     \multirow{2}{*}{\centering\textbf{Avg}}       \\
 &  & Gauss.       & Shot       & Impul.     & Defoc.     & Glass      & Motion      & Snow       & Frost      & Fog        & Brit.      & Contr.    & Pixel       & Sat.         \\

\toprule[1pt]
\multicolumn{1}{c|}{MonoFlex}                                &       CVPR'21
&12.97 & 20.42 & 15.02 & 20.37 & 36.51 & 11.61 & 32.26 & 30.61 & 19.69 & \textbf{45.33} & 20.01 & 29.09 & 42.44 & 25.87\\
\multicolumn{1}{l|}{\quad$\bullet$ TENT}                                &       ICLR'21
&29.84 & 38.55 & 34.54 & 34.93 & 40.52 & 25.08 & 39.68 & \underline{40.53} & 40.30 & 44.37 & 44.04 & 41.10 & 43.92 & 38.26\\ 
\multicolumn{1}{l|}{\quad$\bullet$ EATA}                                &       ICML'22
&30.13 & 38.69 & 34.77 & 35.43 & 40.16 & 27.93 & 39.85 & 40.38 & 40.60 & 44.81 & 44.43 & 41.39 & 44.30 & 38.68\\ 
\multicolumn{1}{l|}{\quad$\bullet$ DeYO}                                &       ICLR'24
&30.58 & 38.82 & 34.93 & 36.04 & \textbf{41.00} & 28.64 & \textbf{39.96} & 40.51 & 40.62 & 44.79 & \underline{44.46} & 41.48 & \textbf{44.90} & 38.98\\
\multicolumn{1}{l|}{\quad$\bullet$ MonoTTA}                                &       ECCV'24
&\textbf{32.34} & \underline{39.05} & \underline{35.68} & \underline{36.58} & 40.69 & \underline{30.25} & 39.70 & 40.01 & \underline{41.22} & 44.76 & \textbf{44.88} & \underline{41.91} & 44.20 & \underline{39.33}\\ 

\rowcolor[HTML]{E6F1FF}
\multicolumn{1}{l|}{\quad$\bullet$ Ours}                                &       This paper
&\underline{32.28} & \textbf{39.42} & \textbf{36.78} & \textbf{36.87} & \underline{40.91} & \textbf{31.33} & \underline{39.88} & \textbf{40.71} & \textbf{41.23} & \underline{44.90} & 44.41 & \textbf{42.47} & \underline{44.44} & \textbf{39.66}\\
\toprule[1pt]
\multicolumn{1}{c|}{MonoGround}                                &       CVPR'22
&13.05          & 22.05          & 19.41          & 20.75          & 38.72          & 8.40           & 30.65          & 27.66          & 14.56          & 46.22          & 14.95          & 33.40          & 36.29          & 25.08\\
\multicolumn{1}{l|}{\quad$\bullet$ TENT}                                &       ICLR'21
&34.94          & 42.76          & 37.93          & 37.79          & \underline{44.95}    & 25.15          & \underline{40.67}    & \underline{42.77}    & 41.26          & \textbf{47.05} & 45.12          & \underline{43.73}    & \underline{46.56}    & 40.82\\ 
\multicolumn{1}{l|}{\quad$\bullet$ EATA}                                &       ICML'22
&35.36          & 42.47          & 38.85          & 38.24          & 44.87          & 26.44          & 40.64          & 42.61          & 41.65          & \underline{46.94}    & 45.18          & 43.71          & 46.54          & 41.04\\ 
\multicolumn{1}{l|}{\quad$\bullet$ DeYO}                                &       ICLR'24
&35.88          & 42.07          & \underline{39.86}    & 38.51          & 44.81          & 28.01          & 40.60          & 42.49          & 41.95          & 46.83          & \underline{45.25}    & 43.67          & 46.54          & 41.27\\
\multicolumn{1}{l|}{\quad$\bullet$ MonoTTA}                                &       ECCV'24
&\underline{37.05}    & \underline{42.86}    & 39.52          & \underline{39.25}    & 44.59          & \underline{32.66}    & 40.54          & 42.47          & \underline{42.13}    & 45.95          & 44.98          & 43.38          & 46.15          & \underline{41.66}\\ 

\rowcolor[HTML]{E6F1FF}
\multicolumn{1}{l|}{\quad$\bullet$ Ours}                                &       This paper
&\textbf{37.25} & \textbf{43.31} & \textbf{40.21} & \textbf{39.80} & \textbf{45.30} & \textbf{34.16} & \textbf{41.39} & \textbf{42.80} & \textbf{42.84} & 46.61          & \textbf{45.93} & \textbf{43.80} & \textbf{46.66} & \textbf{42.31}\\ 

\bottomrule[1pt]
\end{tabular}
}
}
\label{tab:level1}
\end{table*}

\begin{table*}[t]
\centering
\vspace{-3.5mm}
\caption{Comparisons with state-of-the-art methods on the KITTI-C \emph{validation} set (severity level 5) with MonoGround. We highlight the best and second results with {\bf bold} and \underline{underline} respectively.}
\vspace{-1mm}
\resizebox{\textwidth}{!}{
{
\fontsize{20}{25}\selectfont
\begin{tabular}{c|c|ccc|ccc|cccc|ccc|c}
\toprule[1pt]
\multicolumn{16}{c}{Car Category}\\
\toprule[1pt]
\multirow{2}{*}{\centering\textbf{Method}}  & \multirow{2}{*}{\centering\textbf{Reference}}   & \multicolumn{3}{c|}{Noise}              & \multicolumn{3}{c|}{Blur}                          & \multicolumn{4}{c|}{Weather}                       & \multicolumn{3}{c|}{Digital}                       &     \multirow{2}{*}{\centering\textbf{Avg}}       \\
 &  & Gauss.       & Shot       & Impul.     & Defoc.     & Glass      & Motion      & Snow       & Frost      & Fog        & Brit.      & Contr.    & Pixel       & Sat.         \\

\toprule[1pt]
\multicolumn{1}{c|}{MonoGround}                                &       CVPR'22
&0.00 & 0.00  & 0.00  & 0.00 & 11.63 & 0.29 & 1.95  & 6.59  & 3.14  & 19.25 & 0.00  & 4.66  & 3.74  & 3.94\\
\multicolumn{1}{l|}{\quad$\bullet$ TENT}                                &       ICLR'21
&6.82 & 14.81 & 8.21  & 4.88 & 28.38 & 2.65 & 23.92 & 28.08 & 33.06 & 36.70 & 20.22 & 30.63 & 33.27 & 20.90\\ 
\multicolumn{1}{l|}{\quad$\bullet$ EATA}                                &       ICML'22
&7.12 & 15.26 & 8.81  & 5.09 & 29.08 & 2.52 & 24.18 & 28.03 & 33.43 & 36.78 & 21.61 & 30.50 & 33.42 & 21.22\\ 
\multicolumn{1}{l|}{\quad$\bullet$ DeYO}                                &       ICLR'24
&7.35 & 15.72 & 9.38  & 5.74 & 30.01 & 2.99 & 25.03 & 28.55 & 34.32 & 37.31 & 23.41 & 30.99 & 34.16 & 21.92\\
\multicolumn{1}{l|}{\quad$\bullet$ MonoTTA}                                &       ECCV'24
&\underline{7.88} & \underline{16.73} & \underline{10.35} & \underline{5.97} & \underline{31.19} & \underline{3.06} & \underline{25.24} & \underline{28.99} & \underline{34.85} & \underline{37.82} & \underline{25.00} & \underline{31.61} & \underline{34.79} & \underline{22.57}\\ 

\rowcolor[HTML]{E6F1FF}
\multicolumn{1}{l|}{\quad$\bullet$ Ours}                                &       This paper
&\textbf{9.72} & \textbf{18.88} & \textbf{12.74} & \textbf{7.24} & \textbf{33.02} & \textbf{5.24} & \textbf{28.50} & \textbf{30.73} & \textbf{37.27} & \textbf{39.40} & \textbf{28.34} & \textbf{33.22} & \textbf{37.24} & \textbf{24.73}\\

\toprule[1pt]
\multicolumn{16}{c}{Pedestrian Category}\\
\toprule[1pt]
\multirow{2}{*}{\centering\textbf{Method}}  & \multirow{2}{*}{\centering\textbf{Reference}}   & \multicolumn{3}{c|}{Noise}              & \multicolumn{3}{c|}{Blur}                          & \multicolumn{4}{c|}{Weather}                       & \multicolumn{3}{c|}{Digital}                       &     \multirow{2}{*}{\centering\textbf{Avg}}       \\
 &  & Gauss.       & Shot       & Impul.     & Defoc.     & Glass      & Motion      & Snow       & Frost      & Fog        & Brit.      & Contr.    & Pixel       & Sat.         \\

\toprule[1pt]
\multicolumn{1}{c|}{MonoGround}                                &       CVPR'22
&0.00 & 0.00 & 0.00 & 0.00 & 14.76 & 0.00 & 0.28  & 0.74  & 0.68  & 4.63  & 0.00  & 0.34  & 1.80  & 1.79\\
\multicolumn{1}{l|}{\quad$\bullet$ TENT}                                &       ICLR'21
&1.47 & 2.91 & 1.01 & 1.19 & 15.19 & 0.66 & 6.98  & 10.44 & \underline{14.95} & 17.49 & 11.10 & 10.72 & 8.72  & 7.91\\ 
\multicolumn{1}{l|}{\quad$\bullet$ EATA}                                &       ICML'22
&1.85 & 2.86 & 1.05 & 1.31 & 14.02 & 0.79 & 7.41  & 10.08 & 14.72 & 17.57 & 11.31 & 11.20 & 9.38  & 7.97\\ 
\multicolumn{1}{l|}{\quad$\bullet$ DeYO}                                &       ICLR'24
&2.25 & 2.81 & 1.08 & 1.46 & 13.28 & 0.92 & 7.75  & 9.74  & 14.45 & 17.64 & \underline{11.49} & 11.68 & 9.99  & 8.04\\
\multicolumn{1}{l|}{\quad$\bullet$ MonoTTA}                                &       ECCV'24
&\textbf{2.40} & \underline{4.74} & \underline{1.52} & \underline{1.60} & \textbf{16.31} & \underline{1.09} & \underline{8.95}  & \underline{11.06} & 14.72 & \underline{17.96} & 10.62 & \textbf{12.39} & \underline{12.11} & \underline{8.88}\\ 

\rowcolor[HTML]{E6F1FF}
\multicolumn{1}{l|}{\quad$\bullet$ Ours}                                &       This paper
&\underline{2.26} & \textbf{5.03} & \textbf{1.85} & \textbf{2.24} & \underline{16.26} & \textbf{2.29} & \textbf{10.44} & \textbf{12.37} & \textbf{15.50} & \textbf{18.89} & \textbf{12.59} & \underline{12.35} & \textbf{12.95} & \textbf{9.62}\\ 

\toprule[1pt]
\multicolumn{16}{c}{Cyclist Category}\\
\toprule[1pt]
\multirow{2}{*}{\centering\textbf{Method}}  & \multirow{2}{*}{\centering\textbf{Reference}}   & \multicolumn{3}{c|}{Noise}              & \multicolumn{3}{c|}{Blur}                          & \multicolumn{4}{c|}{Weather}                       & \multicolumn{3}{c|}{Digital}                       &     \multirow{2}{*}{\centering\textbf{Avg}}       \\
 &  & Gauss.       & Shot       & Impul.     & Defoc.     & Glass      & Motion      & Snow       & Frost      & Fog        & Brit.      & Contr.    & Pixel       & Sat.         \\

\toprule[1pt]
\multicolumn{1}{c|}{MonoGround}                                &       CVPR'22
&0.00 & 0.00 & 0.00 & 0.00 & 0.47 & 0.00 & 0.10 & 1.20 & 0.21 & 3.85 & 0.00 & 0.76 & 0.19 & 0.52\\
\multicolumn{1}{l|}{\quad$\bullet$ TENT}                                &       ICLR'21
&\textbf{1.77} & \textbf{0.14} & 0.04          & 0.07          & 2.92          & 0.31          & 1.92          & 2.70          & 6.90          & 8.14          & 1.08          & 1.51          & 2.71          & 2.32\\ 
\multicolumn{1}{l|}{\quad$\bullet$ EATA}                                &       ICML'22
&\underline{0.88}    & 0.13          & 0.05          & 0.06          & 2.94          & 0.40          & 2.03          & 2.81          & 6.91          & \underline{8.36}    & 1.32          & 1.52          & 2.98          & 2.34\\ 
\multicolumn{1}{l|}{\quad$\bullet$ DeYO}                                &       ICLR'24
&0.00          & 0.12          & \textbf{0.07} & 0.06          & 2.96          & 0.49          & 2.13          & 2.93          & 6.91          & \textbf{8.58} & 1.57          & 1.54          & 3.25          & 2.35\\
\multicolumn{1}{l|}{\quad$\bullet$ MonoTTA}                                &       ECCV'24
&0.04          & 0.10          & 0.04          & \underline{0.15}    & \underline{3.59}    & \underline{0.52}    & \underline{2.51}    & 3.96          & \underline{8.45}    & 7.80          & \underline{3.00}    & \textbf{2.90} & \underline{3.61}    & \underline{2.82}\\ 

\rowcolor[HTML]{E6F1FF}
\multicolumn{1}{l|}{\quad$\bullet$ Ours}                                &       This paper
&0.05          & \textbf{0.14} & \textbf{0.07} & \textbf{0.24} & \textbf{4.01} & \textbf{0.70} & \textbf{2.67} & \textbf{4.20} & \textbf{8.79} & 8.22          & \textbf{3.91} & \underline{ 2.55}    & \textbf{3.72} & \textbf{3.02}\\

\bottomrule[1pt]
\end{tabular}
}
}
\label{tab:complete_monoground}
\vspace{-5.5mm}
\end{table*}

\section{Additional Ablation Study}
\label{sm:ablation}
In Sec. 6.3 of the main paper, we provide a comprehensive analysis of each component's effectiveness and their complementary interactions. In this section, we extend our analysis by examining the sensitivity of key parameters and comparing running times, offering further insights into the robustness and efficiency of our method.
\subsection{Hyperparameter Robustness}

\begin{figure}[!h]
\begin{center}
\includegraphics[width=0.7\linewidth]{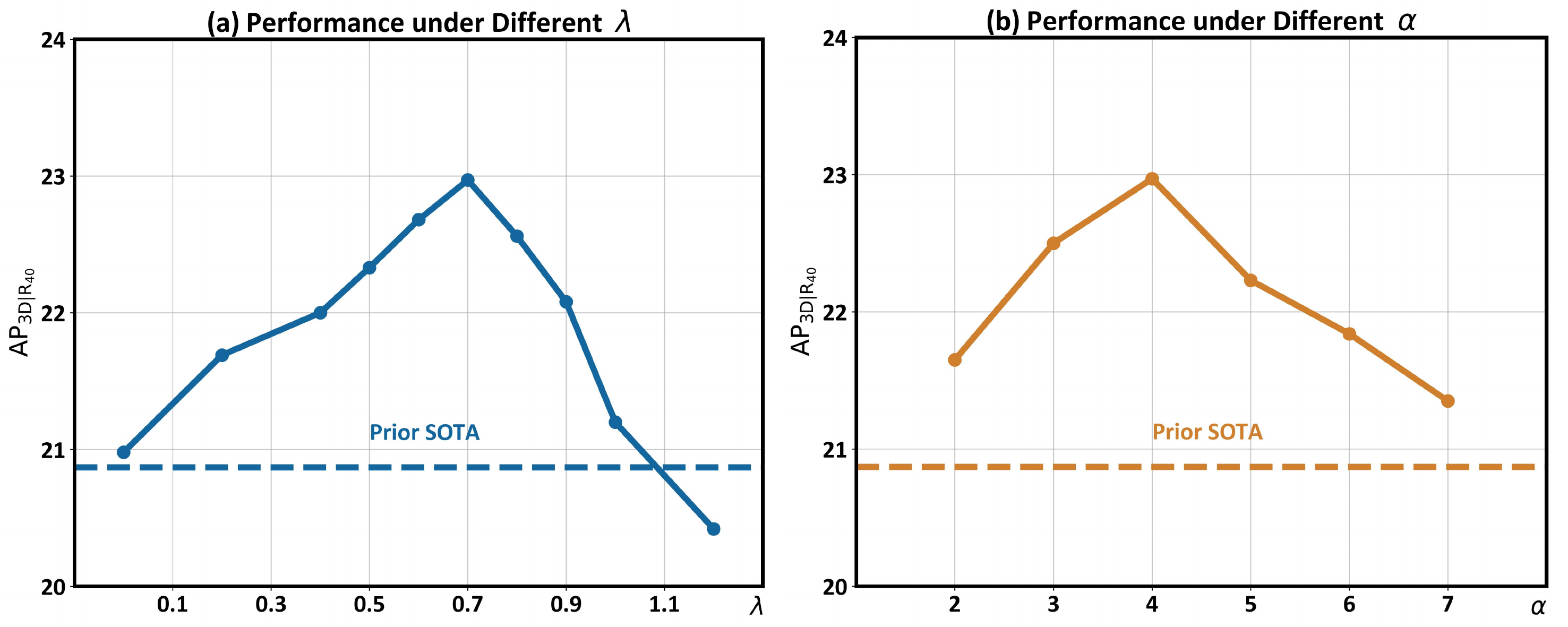}
\end{center}
\caption{({\it a}) Performance with varying strengths $\lambda$ of the normal field constraint. ({\it b}) Performance with different weighting scare $\alpha$ of the conjugate focal loss.} 
\label{fig:hyper}
\end{figure}

Our method involves two key hyperparameters: the coefficient $\lambda$, which determines the trade-off of the semantic and geometric uncertainty optimization, and the coefficient $\alpha$, which controls the weighting scale in conjugate focal loss. We conduct ablation experiments on these two key coefficients independently:

As shown in Fig. \ref{fig:hyper}(a), the different strengths of geometric constant yields stable performance gains. However, when $\lambda$ exceeds the optimal range (\eg, 1.1), the model tends to over-prioritize geometric consistency over uncertainty optimization, leading to worse performance. To balance the effects of two components in our method, we set $\lambda$ to 0.7 by default. In Fig. \ref{fig:hyper}(b), we observe that the weighting coefficient $\alpha$ consistently outperforms prior SOTA methods, demonstrating the robustness of our Conjugate Focal Loss weighting scheme. Empirically, we set $\alpha$ to 4 by default.

Notably, the default choice of $\alpha$ and $\gamma$ not only yields strong empirical performance but also aligns with the standard settings used in vanilla focal loss during source training. This compatibility echoes our theoretical analysis in Sec.\ref{sec:5.1}, suggesting that hyperparameters can remain unchanged from the source phase. Such consistency removes the need for extensive hyperparameter tuning, significantly improving the efficiency and practicality of our adaptation strategy.

\subsection{Running Time Comparison}
In our experiments, we have demonstrated the effectiveness of DUO in various scenarios. In this subsection, we focus on the computational efficiency of DUO. Although our method relies on dual-branch optimization, the computation of the geometric constraint with efficient operators incurs only a slight time cost. As shown in Table \ref{table:running time}, DUO’s running time is less than half that of DeYO, which relies heavily on data augmentation to optimize the uncertainty. Moreover, DUO's adaptation efficiency exceeds that of MonoTTA, which only optimizes semantic uncertainty without addressing geometric uncertainty. Notably, processing 1k images with DUO adds only an extra 6 seconds compared to inference alone, underscoring the high efficiency of our dual-branch optimization framework.

\begin{table}[h]
\begin{center}
\caption{Running time comparison of various methods. We assess TTA approaches for processing 1k images in Gaussian corruption type, using a single Nvidia RTX 4090 GPU.}
\label{table:running time}
\resizebox{0.6\linewidth}{!}{
\begin{tabular}{c|ccccc|c}
\toprule[1pt]
Metrics & Source Model &  TENT & EATA & DeYO  & MonoTTA   & Ours \\ \hline
Running Time  & 26s  & 31s & 29s  & 87s  &  33s   & 32s \\
\toprule[1pt]
\end{tabular}}
\end{center}

\end{table}
\section{More Implementation Details}
\label{sm:implementation}
\subsection{Baseline Methods}
We compare our DUO with several state-of-the-art methods. TENT \citep{tent} reduces the entropy of test samples to guide model updates, prompting the model to generate more confident predictions. Building on this, EATA \citep{eata} incorporates a sample selection mechanism based on low uncertainty to specifically minimize entropy for the most reliable samples, thereby further reducing semantic uncertainty. DeYO \citep{deyo} prioritizes samples with dominant shape information and employs a dual semantic uncertainty criterion to identify reliable samples for adaptation. MonoTTA \citep{lin2025monotta} introduces a negative regularization term on low-score objects, leveraging their negative class information to reduce uncertainty.
\subsection{Detailed Model Architecture}
\label{sm:model}
Our framework is built on a widely-adopted multi-branch architecture for monocular 3D object detection, where separate branches predict various object properties to simultaneously achieve recognition and spatial localization. In 3D detection, accurate depth estimation is a critical factor that significantly influences overall performance \cite{Ma_2021_CVPR}. To enhance depth prediction, many existing models adopt a multi-head strategy that integrates diverse depth estimates to reduce the individual bias. For example, MonoFlex \cite{zhang2021objects} combines direct regression with multiple keypoint estimation; MonoGround leverages ground plane priors for refined depth predictions \cite{qin2022monoground}; and MonoCD exploits the complementary strengths of multiple prediction heads \cite{yan2024monocd}.

To effectively integrate the multi-head predictions, these models includes an uncertainty estimation branch that quantifies the reliability of each depth prediction. The final depth estimation is computed as an uncertainty-weighted average, as shown in the following formulation: 
\begin{equation}
z_{\text {soft }}=\left(\sum_{i=1}^{n} \frac{z_i}{\sigma_i}\right) /\left(\sum_{i=1}^{n} \frac{1}{\sigma_i}\right),
\end{equation}
where the $\sigma_i$ is the uncertainty of the corresponding depth estimation and $n$ is the number of depth heads. In the main paper, we use the average of the $\log\sigma_i$ as the \textbf{depth uncertainty metric}.

Furthermore, the uncertainty regression loss for the entire depth branch is designed as:
\begin{equation}
L_{dep}=\sum_{i}\left[\frac{\left|z_i-z^*\right|}{\sigma_i}+ \log \left(\sigma_i\right)\right],
\end{equation}
where the $z^*$ is the ground-truth depth. 

For our TTA setting, we attempt to utilize the weighted average $z_{soft}$ as a pseudo-label to optimize this loss, directly optimizing the depth uncertainties. However, this approach can lead to model collapse—a phenomenon we analyze in detail in Sec. 4 of the main paper.

For Monoflex~\cite{zhang2021objects} and MonoGround~\cite{qin2022monoground}, we follow their original settings by using a randomly generated seed. Both Monoflex and MonoGround employ the same modified DLA-34~\cite{yu2018deep} as their backbone network, with input resolutions of $384 \times 1280$ for the KITTI-C and $928 \times 1600$ for nuScenes, respectively.

\subsection{More Details on Dataset}
\label{sm:data}
\begin{figure}[!h]
    \centering
    \includegraphics[width=\textwidth]{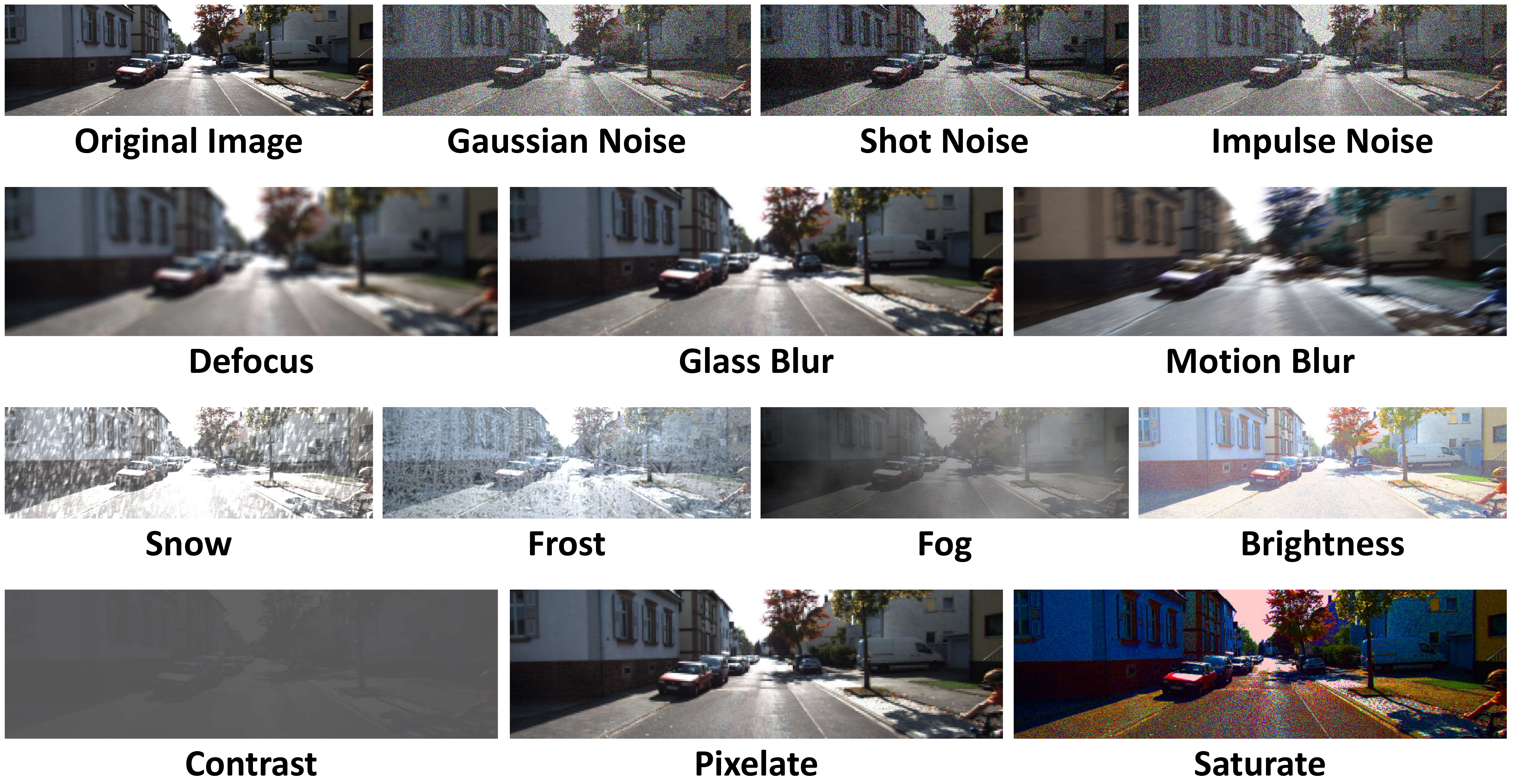}
    \caption{An illustration of 13 distinct types of corruptions in the severity level 3 of the KITTI-C dataset.}
    \label{fig:kitti}
\end{figure}

\noindent\textbf{KITTI-C Dataset.} We follow the protocol from \cite{zhang2021objects,lin2025monotta} to partition the KITTI dataset into a training set (3712 images) and a validation set (3769 images) for model training and adaptation, respectively. For evaluation, we employ the KITTI-C version, which applies 13 distinct corruptions to the validation set—namely, Gaussian noise, shot noise, impulse noise, defocus blur, glass blur, motion blur, snow, frost, fog, brightness, contrast, pixelation, and saturation \cite{hendrycks2018benchmarking}, as shown in Fig. \ref{fig:kitti}. Each corruption is further divided into five severity levels, with higher levels indicating more extreme perturbations and distribution shifts.

\begin{figure}[!h]
    \centering
    \includegraphics[width=\textwidth]{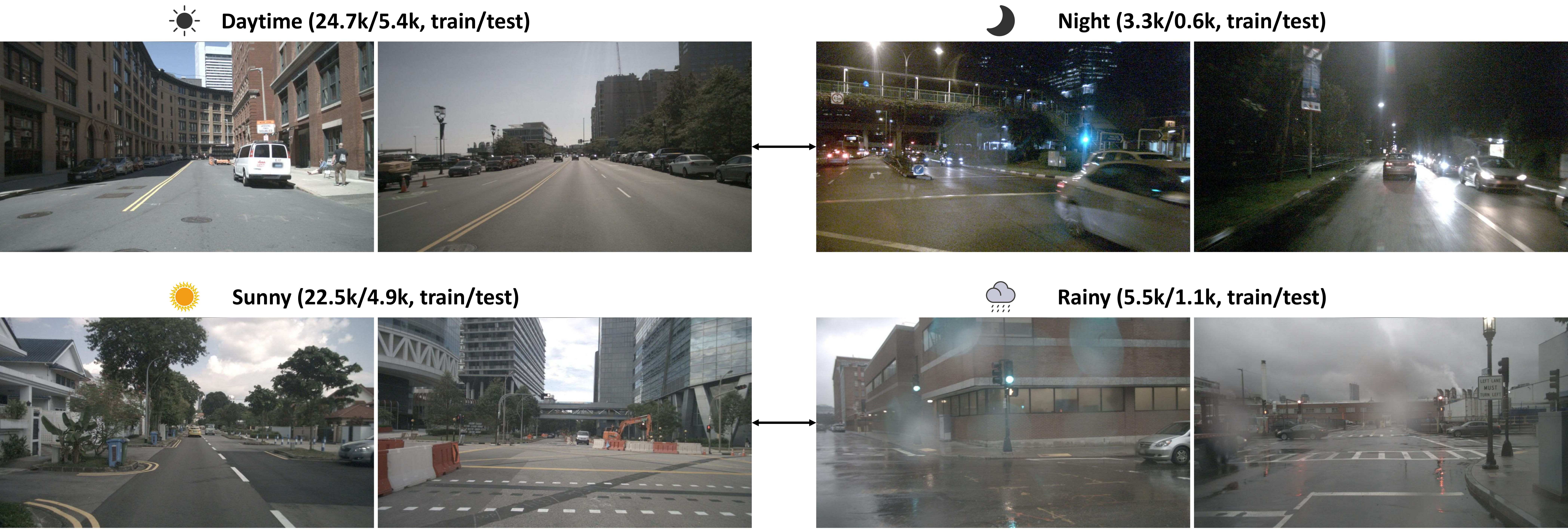}
    \caption{An illustration of the Daytime, Night, Sunny, and Rainy scenarios of the nuScenes dataset.}
    \label{fig:nuscene}
\end{figure}

\noindent\textbf{nuScenes Dataset.}
For the four real-world scenarios in the nuScenes dataset, we first extract all \emph{front-view} images and convert them to KITTI format using the official devkit~\cite{caesar2020nuscenes}. Following \cite{liu2023bevfusion}, we partition these images into Daytime, Night, Sunny, and Rainy scenarios based on their scene descriptions. For each scenario, we train our model on the training split and evaluate its performance on the validation split (the number of images per scenario is shown in Fig. \ref{fig:nuscene}). Since the Night scenario contains fewer than 4k images with fewer objects (e.g., pedestrians), we report results only for the \emph{Car} category.

%% file: main.bbl
\begin{thebibliography}{56}
\providecommand{\natexlab}[1]{#1}
\providecommand{\url}[1]{\texttt{#1}}
\expandafter\ifx\csname urlstyle\endcsname\relax
  \providecommand{\doi}[1]{doi: #1}\else
  \providecommand{\doi}{doi: \begingroup \urlstyle{rm}\Url}\fi

\bibitem[Arnold et~al.(2019)Arnold, Al-Jarrah, Dianati, Fallah, Oxtoby, and Mouzakitis]{m3od2}
Eduardo Arnold, Omar~Y Al-Jarrah, Mehrdad Dianati, Saber Fallah, David Oxtoby, and Alex Mouzakitis.
\newblock A survey on 3d object detection methods for autonomous driving applications.
\newblock \emph{IEEE Transactions on Intelligent Transportation Systems}, 2019.

\bibitem[Ben-David et~al.(2010)Ben-David, Blitzer, Crammer, Kulesza, Pereira, and Vaughan]{ben2010theory}
Shai Ben-David, John Blitzer, Koby Crammer, Alex Kulesza, Fernando Pereira, and Jennifer~Wortman Vaughan.
\newblock A theory of learning from different domains.
\newblock \emph{Machine learning}, 79:\penalty0 151--175, 2010.

\bibitem[Bertsekas et~al.(2003)Bertsekas, Nedic, and Ozdaglar]{bertsekas2003convex}
Dimitri Bertsekas, Angelia Nedic, and Asuman Ozdaglar.
\newblock \emph{Convex analysis and optimization}.
\newblock Athena Scientific, 2003.

\bibitem[Boudiaf et~al.(2022)Boudiaf, Mueller, Ben~Ayed, and Bertinetto]{boudiaf2022parameter}
Malik Boudiaf, Romain Mueller, Ismail Ben~Ayed, and Luca Bertinetto.
\newblock Parameter-free online test-time adaptation.
\newblock In \emph{Proceedings of the IEEE/CVF Conference on Computer Vision and Pattern Recognition}, pages 8344--8353, 2022.

\bibitem[Caesar et~al.(2020)Caesar, Bankiti, Lang, Vora, Liong, Xu, Krishnan, Pan, Baldan, and Beijbom]{caesar2020nuscenes}
Holger Caesar, Varun Bankiti, Alex~H Lang, Sourabh Vora, Venice~Erin Liong, Qiang Xu, Anush Krishnan, Yu Pan, Giancarlo Baldan, and Oscar Beijbom.
\newblock nuscenes: A multimodal dataset for autonomous driving.
\newblock In \emph{Proceedings of the IEEE/CVF conference on computer vision and pattern recognition}, pages 11621--11631, 2020.

\bibitem[Chen et~al.(2020{\natexlab{a}})Chen, Wu, Liu, and Xu]{chen2020foreground}
Joya Chen, Qi Wu, Dong Liu, and Tong Xu.
\newblock Foreground-background imbalance problem in deep object detectors: A review.
\newblock In \emph{2020 IEEE Conference on Multimedia Information Processing and Retrieval (MIPR)}, pages 285--290. IEEE, 2020{\natexlab{a}}.

\bibitem[Chen et~al.(2020{\natexlab{b}})Chen, Tai, Sun, and Li]{chen2020monopair}
Yongjian Chen, Lei Tai, Kai Sun, and Mingyang Li.
\newblock Monopair: Monocular 3d object detection using pairwise spatial relationships.
\newblock In \emph{Proceedings of the IEEE/CVF Conference on Computer Vision and Pattern Recognition}, pages 12093--12102, 2020{\natexlab{b}}.

\bibitem[Ding et~al.(2020)Ding, Huo, Yi, Wang, Shi, Lu, and Luo]{ding2020learning}
Mingyu Ding, Yuqi Huo, Hongwei Yi, Zhe Wang, Jianping Shi, Zhiwu Lu, and Ping Luo.
\newblock Learning depth-guided convolutions for monocular 3d object detection.
\newblock In \emph{Proceedings of the IEEE/CVF Conference on computer vision and pattern recognition workshops}, pages 1000--1001, 2020.

\bibitem[Erhan et~al.(2010)Erhan, Courville, Bengio, and Vincent]{erhan2010does}
Dumitru Erhan, Aaron Courville, Yoshua Bengio, and Pascal Vincent.
\newblock Why does unsupervised pre-training help deep learning?
\newblock In \emph{Proceedings of the thirteenth international conference on artificial intelligence and statistics}, pages 201--208. JMLR Workshop and Conference Proceedings, 2010.

\bibitem[Gao et~al.(2024)Gao, Yao, and Xu]{gaofast}
Junyu Gao, Xuan Yao, and Changsheng Xu.
\newblock Fast-slow test-time adaptation for online vision-and-language navigation.
\newblock In \emph{Forty-first International Conference on Machine Learning}, 2024.

\bibitem[Geiger et~al.(2012)Geiger, Lenz, and Urtasun]{geiger2012we}
Andreas Geiger, Philip Lenz, and Raquel Urtasun.
\newblock Are we ready for autonomous driving? the kitti vision benchmark suite.
\newblock In \emph{Proceedings of the IEEE/CVF Conference on Computer Vision and Pattern Recognition}, pages 3354--3361. IEEE, 2012.

\bibitem[Goyal et~al.(2022)Goyal, Sun, Raghunathan, and Kolter]{goyal2022test}
Sachin Goyal, Mingjie Sun, Aditi Raghunathan, and J~Zico Kolter.
\newblock Test time adaptation via conjugate pseudo-labels.
\newblock \emph{Advances in Neural Information Processing Systems}, 35:\penalty0 6204--6218, 2022.

\bibitem[Hakim et~al.(2023)Hakim, Osowiechi, Noori, Cheraghalikhani, Bahri, Ben~Ayed, and Desrosiers]{hakim2023clust3}
Gustavo A~Vargas Hakim, David Osowiechi, Mehrdad Noori, Milad Cheraghalikhani, Ali Bahri, Ismail Ben~Ayed, and Christian Desrosiers.
\newblock Clust3: Information invariant test-time training.
\newblock In \emph{Proceedings of the IEEE/CVF International Conference on Computer Vision}, pages 6136--6145, 2023.

\bibitem[Hendrycks and Dietterich(2018)]{hendrycks2018benchmarking}
Dan Hendrycks and Thomas Dietterich.
\newblock Benchmarking neural network robustness to common corruptions and perturbations.
\newblock In \emph{International Conference on Learning Representations}, 2018.

\bibitem[Hornauer and Belagiannis(2022)]{hornauer2022gradient}
Julia Hornauer and Vasileios Belagiannis.
\newblock Gradient-based uncertainty for monocular depth estimation.
\newblock In \emph{Proceedings of the European Conference on Computer Vision}, pages 613--630. Springer, 2022.

\bibitem[Hu et~al.(2025{\natexlab{a}})Hu, Hu, and Duan]{hu2025seva}
Zixuan Hu, Yichun Hu, and Ling-Yu Duan.
\newblock Seva: Leveraging single-step ensemble of vicinal augmentations for test-time adaptation.
\newblock \emph{arXiv preprint arXiv:2505.04087}, 2025{\natexlab{a}}.

\bibitem[Hu et~al.(2025{\natexlab{b}})Hu, Hu, Li, Tang, and Duan]{hu2025beyond}
Zixuan Hu, Yichun Hu, Xiaotong Li, Shixiang Tang, and Ling-Yu Duan.
\newblock Beyond entropy: Region confidence proxy for wild test-time adaptation.
\newblock \emph{arXiv preprint arXiv:2505.20704}, 2025{\natexlab{b}}.

\bibitem[Huang et~al.(2022)Huang, Wu, Su, and Hsu]{huang2022monodtr}
Kuan-Chih Huang, Tsung-Han Wu, Hung-Ting Su, and Winston~H Hsu.
\newblock Monodtr: Monocular 3d object detection with depth-aware transformer.
\newblock In \emph{Proceedings of the IEEE/CVF conference on computer vision and pattern recognition}, pages 4012--4021, 2022.

\bibitem[Iwasawa and Matsuo(2021)]{iwasawa2021test}
Yusuke Iwasawa and Yutaka Matsuo.
\newblock Test-time classifier adjustment module for model-agnostic domain generalization.
\newblock \emph{Advances in Neural Information Processing Systems}, 34:\penalty0 2427--2440, 2021.

\bibitem[Kanopoulos et~al.(1988)Kanopoulos, Vasanthavada, and Baker]{kanopoulos1988design}
Nick Kanopoulos, Nagesh Vasanthavada, and Robert~L Baker.
\newblock Design of an image edge detection filter using the sobel operator.
\newblock \emph{IEEE Journal of solid-state circuits}, 23\penalty0 (2):\penalty0 358--367, 1988.

\bibitem[Karmanov et~al.(2024)Karmanov, Guan, Lu, El~Saddik, and Xing]{karmanov2024efficient}
Adilbek Karmanov, Dayan Guan, Shijian Lu, Abdulmotaleb El~Saddik, and Eric Xing.
\newblock Efficient test-time adaptation of vision-language models.
\newblock In \emph{Proceedings of the IEEE/CVF Conference on Computer Vision and Pattern Recognition}, pages 14162--14171, 2024.

\bibitem[Lee et~al.(2024)Lee, Jung, Lee, Park, Shin, Hwang, and Yoon]{deyo}
Jonghyun Lee, Dahuin Jung, Saehyung Lee, Junsung Park, Juhyeon Shin, Uiwon Hwang, and Sungroh Yoon.
\newblock Entropy is not enough for test-time adaptation: From the perspective of disentangled factors.
\newblock In \emph{The Twelfth International Conference on Learning Representations}, 2024.

\bibitem[Liang et~al.(2024)Liang, He, and Tan]{liang2024comprehensive}
Jian Liang, Ran He, and Tieniu Tan.
\newblock A comprehensive survey on test-time adaptation under distribution shifts.
\newblock \emph{International Journal of Computer Vision}, pages 1--34, 2024.

\bibitem[Lin et~al.(2025)Lin, Zhang, Niu, Cui, and Li]{lin2025monotta}
Hongbin Lin, Yifan Zhang, Shuaicheng Niu, Shuguang Cui, and Zhen Li.
\newblock Monotta: Fully test-time adaptation for monocular 3d object detection.
\newblock In \emph{Proceedings of the European Conference on Computer Vision}, pages 96--114. Springer, 2025.

\bibitem[Lin et~al.(2017)Lin, Goyal, Girshick, He, and Doll{\'a}r]{lin2017focal}
Tsung-Yi Lin, Priya Goyal, Ross Girshick, Kaiming He, and Piotr Doll{\'a}r.
\newblock Focal loss for dense object detection.
\newblock In \emph{Proceedings of the IEEE international conference on computer vision}, pages 2980--2988, 2017.

\bibitem[Lin et~al.(2023)Lin, Mirza, Kozinski, Possegger, Kuehne, and Bischof]{lin2023video}
Wei Lin, Muhammad~Jehanzeb Mirza, Mateusz Kozinski, Horst Possegger, Hilde Kuehne, and Horst Bischof.
\newblock Video test-time adaptation for action recognition.
\newblock In \emph{Proceedings of the IEEE/CVF Conference on Computer Vision and Pattern Recognition}, pages 22952--22961, 2023.

\bibitem[Liu et~al.(2024)Liu, Shen, Li, Bi, Liu, Pun, and Cun]{liu2024depth}
Weihuang Liu, Xi Shen, Haolun Li, Xiuli Bi, Bo Liu, Chi-Man Pun, and Xiaodong Cun.
\newblock Depth-aware test-time training for zero-shot video object segmentation.
\newblock In \emph{Proceedings of the IEEE/CVF Conference on Computer Vision and Pattern Recognition}, pages 19218--19227, 2024.

\bibitem[Liu et~al.(2021)Liu, Zhou, Lu, Fang, and Zhang]{liu2021autoshape}
Zongdai Liu, Dingfu Zhou, Feixiang Lu, Jin Fang, and Liangjun Zhang.
\newblock Autoshape: Real-time shape-aware monocular 3d object detection.
\newblock In \emph{Proceedings of the IEEE/CVF International Conference on Computer Vision}, pages 15641--15650, 2021.

\bibitem[Liu et~al.(2023)Liu, Tang, Amini, Yang, Mao, Rus, and Han]{liu2023bevfusion}
Zhijian Liu, Haotian Tang, Alexander Amini, Xinyu Yang, Huizi Mao, Daniela~L Rus, and Song Han.
\newblock Bevfusion: Multi-task multi-sensor fusion with unified bird's-eye view representation.
\newblock In \emph{2023 IEEE international conference on robotics and automation (ICRA)}, pages 2774--2781. IEEE, 2023.

\bibitem[Lu et~al.(2021)Lu, Ma, Yang, Zhang, Liu, Chu, Yan, and Ouyang]{lu2021geometry}
Yan Lu, Xinzhu Ma, Lei Yang, Tianzhu Zhang, Yating Liu, Qi Chu, Junjie Yan, and Wanli Ouyang.
\newblock Geometry uncertainty projection network for monocular 3d object detection.
\newblock In \emph{Proceedings of the IEEE/CVF International Conference on Computer Vision}, pages 3111--3121, 2021.

\bibitem[Luo et~al.(2023)Luo, Zheng, Yan, Kun, Zheng, Cui, and Li]{luo2023latr}
Yueru Luo, Chaoda Zheng, Xu Yan, Tang Kun, Chao Zheng, Shuguang Cui, and Zhen Li.
\newblock Latr: 3d lane detection from monocular images with transformer.
\newblock In \emph{Proceedings of the IEEE/CVF International Conference on Computer Vision}, pages 7941--7952, 2023.

\bibitem[Ma et~al.(2021)Ma, Zhang, Xu, Zhou, Yi, Li, and Ouyang]{Ma_2021_CVPR}
Xinzhu Ma, Yinmin Zhang, Dan Xu, Dongzhan Zhou, Shuai Yi, Haojie Li, and Wanli Ouyang.
\newblock Delving into localization errors for monocular 3d object detection.
\newblock In \emph{Proceedings of the IEEE/CVF Conference on Computer Vision and Pattern Recognition}, pages 4721--4730, 2021.

\bibitem[Ma et~al.(2023)Ma, Ouyang, Simonelli, and Ricci]{m3od1}
Xinzhu Ma, Wanli Ouyang, Andrea Simonelli, and Elisa Ricci.
\newblock 3d object detection from images for autonomous driving: a survey.
\newblock \emph{IEEE Transactions on Pattern Analysis and Machine Intelligence}, 2023.

\bibitem[Ma et~al.(2024)Ma, Wang, Bai, Yang, Hou, Wang, Qiao, Yang, and Zhu]{m3od3}
Yuexin Ma, Tai Wang, Xuyang Bai, Huitong Yang, Yuenan Hou, Yaming Wang, Yu Qiao, Ruigang Yang, and Xinge Zhu.
\newblock Vision-centric bev perception: A survey.
\newblock \emph{IEEE Transactions on Pattern Analysis and Machine Intelligence}, 2024.

\bibitem[Mori et~al.(2022)Mori, Ziyin, Liu, and Ueda]{mori2022power}
Takashi Mori, Liu Ziyin, Kangqiao Liu, and Masahito Ueda.
\newblock Power-law escape rate of sgd.
\newblock In \emph{International Conference on Machine Learning}, pages 15959--15975. PMLR, 2022.

\bibitem[Niu et~al.(2022)Niu, Wu, Zhang, Chen, Zheng, Zhao, and Tan]{eata}
Shuaicheng Niu, Jiaxiang Wu, Yifan Zhang, Yaofo Chen, Shijian Zheng, Peilin Zhao, and Mingkui Tan.
\newblock Efficient test-time model adaptation without forgetting.
\newblock In \emph{International conference on machine learning}, pages 16888--16905. PMLR, 2022.

\bibitem[Niu et~al.(2023)Niu, Wu, Zhang, Wen, Chen, Zhao, and Tan]{sar}
Shuaicheng Niu, Jiaxiang Wu, Yifan Zhang, Zhiquan Wen, Yaofo Chen, Peilin Zhao, and Mingkui Tan.
\newblock Towards stable test-time adaptation in dynamic wild world.
\newblock In \emph{Internetional Conference on Learning Representations}, 2023.

\bibitem[Paszke et~al.(2019)Paszke, Gross, Massa, Lerer, Bradbury, Chanan, Killeen, Lin, Gimelshein, Antiga, et~al.]{paszke2019pytorch}
Adam Paszke, Sam Gross, Francisco Massa, Adam Lerer, James Bradbury, Gregory Chanan, Trevor Killeen, Zeming Lin, Natalia Gimelshein, Luca Antiga, et~al.
\newblock Pytorch: An imperative style, high-performance deep learning library.
\newblock In \emph{Advances in neural information processing systems}, 2019.

\bibitem[Poggi et~al.(2020)Poggi, Aleotti, Tosi, and Mattoccia]{poggi2020uncertainty}
Matteo Poggi, Filippo Aleotti, Fabio Tosi, and Stefano Mattoccia.
\newblock On the uncertainty of self-supervised monocular depth estimation.
\newblock In \emph{Proceedings of the IEEE/CVF conference on computer vision and pattern recognition}, pages 3227--3237, 2020.

\bibitem[Qin and Li(2022)]{qin2022monoground}
Zequn Qin and Xi Li.
\newblock Monoground: Detecting monocular 3d objects from the ground.
\newblock In \emph{Proceedings of the IEEE/CVF Conference on Computer Vision and Pattern Recognition}, pages 3793--3802, 2022.

\bibitem[Reading et~al.(2021)Reading, Harakeh, Chae, and Waslander]{reading2021categorical}
Cody Reading, Ali Harakeh, Julia Chae, and Steven~L Waslander.
\newblock Categorical depth distribution network for monocular 3d object detection.
\newblock In \emph{Proceedings of the IEEE/CVF Conference on Computer Vision and Pattern Recognition}, pages 8555--8564, 2021.

\bibitem[S and Fleuret(2021)]{S_2021_CVPR}
Prabhu~Teja S and Francois Fleuret.
\newblock Uncertainty reduction for model adaptation in semantic segmentation.
\newblock In \emph{Proceedings of the IEEE/CVF Conference on Computer Vision and Pattern Recognition}, pages 9613--9623, 2021.

\bibitem[Simonelli et~al.(2019)Simonelli, Bulo, Porzi, L{\'o}pez-Antequera, and Kontschieder]{simonelli2019disentangling}
Andrea Simonelli, Samuel~Rota Bulo, Lorenzo Porzi, Manuel L{\'o}pez-Antequera, and Peter Kontschieder.
\newblock Disentangling monocular 3d object detection.
\newblock In \emph{Proceedings of the IEEE/CVF international conference on computer vision}, pages 1991--1999, 2019.

\bibitem[Sun et~al.(2020)Sun, Wang, Liu, Miller, Efros, and Hardt]{ttt}
Yu Sun, Xiaolong Wang, Zhuang Liu, John Miller, Alexei Efros, and Moritz Hardt.
\newblock Test-time training with self-supervision for generalization under distribution shifts.
\newblock In \emph{International conference on machine learning}, pages 9229--9248. PMLR, 2020.

\bibitem[Touchette(2005)]{touchette2005legendre}
Hugo Touchette.
\newblock Legendre-fenchel transforms in a nutshell.
\newblock \emph{URL http://www. maths. qmul. ac. uk/ht/archive/lfth2. pdf}, page~25, 2005.

\bibitem[Wang et~al.(2020{\natexlab{a}})Wang, Shelhamer, Liu, Olshausen, and Darrell]{tent}
Dequan Wang, Evan Shelhamer, Shaoteng Liu, Bruno Olshausen, and Trevor Darrell.
\newblock Tent: Fully test-time adaptation by entropy minimization.
\newblock In \emph{International Conference on Learning Representations}, 2020{\natexlab{a}}.

\bibitem[Wang et~al.(2021{\natexlab{a}})Wang, Du, Ye, Fu, Guo, Xue, Feng, and Zhang]{wang2021depth}
Li Wang, Liang Du, Xiaoqing Ye, Yanwei Fu, Guodong Guo, Xiangyang Xue, Jianfeng Feng, and Li Zhang.
\newblock Depth-conditioned dynamic message propagation for monocular 3d object detection.
\newblock In \emph{Proceedings of the IEEE/CVF Conference on Computer Vision and Pattern Recognition}, pages 454--463, 2021{\natexlab{a}}.

\bibitem[Wang et~al.(2022)Wang, Xinge, Pang, and Lin]{wang2022probabilistic}
Tai Wang, ZHU Xinge, Jiangmiao Pang, and Dahua Lin.
\newblock Probabilistic and geometric depth: Detecting objects in perspective.
\newblock In \emph{Conference on Robot Learning}, pages 1475--1485. PMLR, 2022.

\bibitem[Wang et~al.(2020{\natexlab{b}})Wang, Chen, You, Li, Hariharan, Campbell, Weinberger, and Chao]{wang2020train}
Yan Wang, Xiangyu Chen, Yurong You, Li~Erran Li, Bharath Hariharan, Mark Campbell, Kilian~Q Weinberger, and Wei-Lun Chao.
\newblock Train in germany, test in the usa: Making 3d object detectors generalize.
\newblock In \emph{Proceedings of the IEEE/CVF Conference on Computer Vision and Pattern Recognition}, pages 11713--11723, 2020{\natexlab{b}}.

\bibitem[Wang et~al.(2021{\natexlab{b}})Wang, Peng, and Zhang]{Wang_2021_ICCV}
Yuxi Wang, Junran Peng, and ZhaoXiang Zhang.
\newblock Uncertainty-aware pseudo label refinery for domain adaptive semantic segmentation.
\newblock In \emph{Proceedings of the IEEE/CVF International Conference on Computer Vision}, pages 9092--9101, 2021{\natexlab{b}}.

\bibitem[Wang et~al.(2024)Wang, Cheraghian, Hayder, Hong, Ramasinghe, Rahman, Ahmedt-Aristizabal, Li, Petersson, and Harandi]{wang2024backpropagation}
Yanshuo Wang, Ali Cheraghian, Zeeshan Hayder, Jie Hong, Sameera Ramasinghe, Shafin Rahman, David Ahmedt-Aristizabal, Xuesong Li, Lars Petersson, and Mehrtash Harandi.
\newblock Backpropagation-free network for 3d test-time adaptation.
\newblock In \emph{Proceedings of the IEEE/CVF Conference on Computer Vision and Pattern Recognition}, pages 23231--23241, 2024.

\bibitem[Xu et~al.(2023)Xu, Peng, Cheng, Li, Qian, Li, Wang, and Cai]{xu2023mononerd}
Junkai Xu, Liang Peng, Haoran Cheng, Hao Li, Wei Qian, Ke Li, Wenxiao Wang, and Deng Cai.
\newblock Mononerd: Nerf-like representations for monocular 3d object detection.
\newblock In \emph{Proceedings of the IEEE/CVF International Conference on Computer Vision}, pages 6814--6824, 2023.

\bibitem[Yan et~al.(2024)Yan, Yan, Xiong, Xiang, and Tan]{yan2024monocd}
Longfei Yan, Pei Yan, Shengzhou Xiong, Xuanyu Xiang, and Yihua Tan.
\newblock Monocd: Monocular 3d object detection with complementary depths.
\newblock In \emph{Proceedings of the IEEE/CVF Conference on Computer Vision and Pattern Recognition}, pages 10248--10257, 2024.

\bibitem[Yu et~al.(2018)Yu, Wang, Shelhamer, and Darrell]{yu2018deep}
Fisher Yu, Dequan Wang, Evan Shelhamer, and Trevor Darrell.
\newblock Deep layer aggregation.
\newblock In \emph{Proceedings of the IEEE conference on computer vision and pattern recognition}, pages 2403--2412, 2018.

\bibitem[Zhang et~al.(2021)Zhang, Lu, and Zhou]{zhang2021objects}
Yunpeng Zhang, Jiwen Lu, and Jie Zhou.
\newblock Objects are different: Flexible monocular 3d object detection.
\newblock In \emph{Proceedings of the IEEE/CVF Conference on Computer Vision and Pattern Recognition}, pages 3289--3298, 2021.

\bibitem[Zhao et~al.(2023)Zhao, Chen, and Xia]{zhao2023delta}
Bowen Zhao, Chen Chen, and Shu-Tao Xia.
\newblock {DELTA}: {DEGRADATION}-{FREE} {FULLY} {TEST}-{TIME} {ADAPTATION}.
\newblock In \emph{The Eleventh International Conference on Learning Representations}, 2023.

\end{thebibliography}
